\documentclass[11pt]{article}

\usepackage[preprint]{acl}

\usepackage{times}
\usepackage{latexsym}

\usepackage[T1]{fontenc}

\usepackage[utf8]{inputenc}

\usepackage{microtype}

\usepackage{inconsolata}

\usepackage{graphicx}

\usepackage{xcolor}
\usepackage{microtype}
\usepackage{hyperref}
\usepackage{url}
\usepackage{booktabs}
\usepackage{amsmath}
\usepackage{amssymb} 
\usepackage{multirow}
\usepackage{subcaption}
\usepackage{verbatim}
\usepackage{cleveref}
\usepackage{tcolorbox}
\usepackage{wrapfig}
\usepackage{tikz}
\usetikzlibrary{positioning, arrows.meta}

\usepackage{lineno}
\usepackage{tikz}
\usetikzlibrary{arrows.meta,positioning,shapes.geometric,shadows,backgrounds,fit}

\definecolor{darkblue}{rgb}{0, 0, 0.5}
\hypersetup{colorlinks=true, citecolor=darkblue, linkcolor=darkblue, urlcolor=darkblue}
\definecolor{posgreen}{RGB}{0,140,0}
\definecolor{negred}{RGB}{180,0,0}

\title{The Count Is There, but Misaligned: Understanding and Correcting Counting Failures in VLMs}

\author{
  \textbf{Ahmed Oumar El-Shangiti\textsuperscript{1,2}} \quad
  \textbf{Abzal Nurgazy\textsuperscript{1}} \quad
  \textbf{Hilal AlQuabeh\textsuperscript{1}}
\\
  \textbf{Nikolai Rozanov\textsuperscript{3}} \quad
  \textbf{Kentaro Inui\textsuperscript{1}}
\\
\\
  \textsuperscript{1}MBZUAI \quad
  \textsuperscript{2}DataBayt.AI Labs\quad
  \textsuperscript{3}Imperial College London
\\
  \small{
    \textbf{Correspondence:} \href{mailto:ahmed.oumar@mbzuai.ac.ae}{ahmed.oumar@mbzuai.ac.ae}
  }
}

\begin{document}
\maketitle
\begin{abstract}
Despite strong performance on many multimodal tasks, vision-language models (VLMs) still struggle with basic object counting. We investigate whether this reflects missing internal knowledge or a gap between internal representations and verbalized outputs. Training simple probes on activations from four VLMs across five counting datasets reveals that nonlinear probes can reliably detect counting errors, suggesting that VLMs often encode the correct count even when they output the wrong answer. SVCCA analysis shows that probes trained on ground-truth counts and probes trained on model outputs occupy a partially shared activation subspace but read out along misaligned directions. We further validate our findings using a causal steering intervention, proving that strengthening the direction of count-identified probes does improve model counting performance. Motivated by this result, we propose a detector-guided self-correction method that selectively re-prompts the model only when an internal error detector predicts failure. This simple inference-time intervention improves counting accuracy by up to $15.6\%$ absolute percentage points, without any parameter updates. Our results establish activation-based error probing as both a practical tool for improving VLM counting and a mechanistic lens on the gap between internal knowledge and model outputs.
\end{abstract}

\section{Introduction}
\label{sec:intro}

Vision-language models (VLMs) have achieved strong performance across a wide range of tasks, including image captioning~\citep{li2023blip2}, visual question answering~\citep{liu2023llava}, reasoning~\citep{lu2024mathvista}, and web navigation~\citep{koh2024visualwebarena}. However, strong performance on these tasks does not imply reliable quantitative perception. Among such capabilities, counting is especially important because it requires a model to detect relevant objects, distinguish them from distractors, maintain consistent correspondences, and map visual evidence to an exact numerical answer. This makes counting a useful test of whether a VLM truly grounds its predictions in the image rather than relying on superficial correlations or language priors~\citep{vo2025vlmbiased}.

At the same time, counting exposes several core failure modes in VLMs, including hallucination, failures under clutter or occlusion, poor object individuation, and mistakes in translating perceptual representations into language. Recent work has shown that these weaknesses remain substantial even in strong contemporary models~\citep{weng2025visnumbench, vo2025vlmbiased, zhang2024countbench, fu2023mme}. 


Most existing work documents that VLMs are weak at counting, but it does so primarily at the behavioral level, through benchmarks and aggregate performance comparisons~\citep{weng2025visnumbench, vo2025vlmbiased, zhang2024countbench, fu2023mme}. These studies are important because they establish counting as a persistent failure mode, yet they largely leave open a more fundamental question: \emph{why} do VLMs fail at counting? In particular, behavioral evaluations can show when a model is wrong, but they do not reveal whether the correct count is absent from the model's internal representations, whether it is present but poorly aligned with the model's eventual answer, or whether the failure emerges later during decoding.

A smaller body of work begins to examine counting more mechanistically~\citep{hasani2025countingmechanism, alghisi2025vlmcounting}, and some methods attempt to improve performance directly. For example, \citet{alghisi2025vlmcounting} retrain parts of the model, while \citet{sengupta2025modest} use attention-based interventions and obtain modest gains. However these approaches primarily target better final behavior or what layers/tokens are involved in the counting process , rather than explaining how count information is internally represented, or why it sometimes fails to appear in the final prediction. 

Our work addresses this gap by analyzing counting through multiple probes applied to intermediate VLM representations, with each probe trained to capture a different part of the model's computation. Rather than asking only whether a model produces the correct count, we ask what information about the ground-truth count, the model's predicted count, and the likelihood of error is present during the forward pass, and how these signals relate to one another across depth. This makes it possible to move beyond behavioral failure and study the internal structure of counting errors.

\begin{figure*}[t]
\centering
\includegraphics[width=\linewidth]{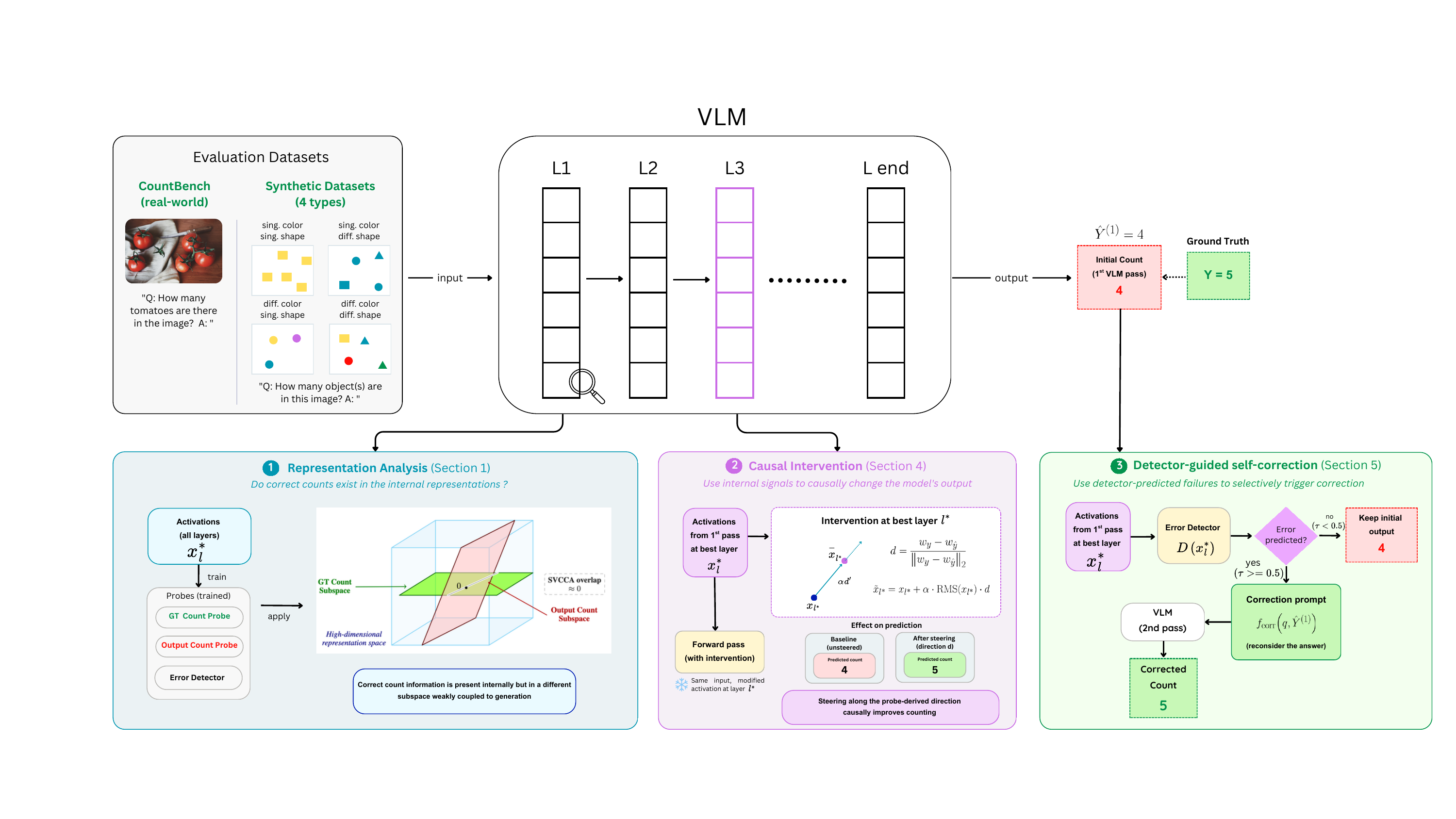}
\caption{Overview of the activation-based probing, intervention, and self-correction framework. The pipeline extracts VLM hidden states to train probes, analyze representational subspaces via SVCCA, causally steer activations along probe-derived count directions, and selectively trigger inference-time correction using an internal error detector.}
\label{fig:pipeline_labeled}
\end{figure*}

Specifically, we extend~\cite{sun2025probarithlang} to VLM counting and train separate probes on the same intermediate representations using different supervision: one to predict the ground-truth count, one to predict the model's own output count, and one to predict whether the model's answer is wrong. This multi-probe setup allows us to move beyond simple decodability and study how different counting-relevant signals are organized internally. Our results suggest that counting errors arise from at least two distinct sources. First, the model's final generated answer does not always faithfully reflect information already present in intermediate representations. Second, even when intermediate representations support the model's own eventual answer, they can remain more strongly aligned with the correct count, revealing a mismatch between internal quantity information and final prediction. To make this comparison explicit, we analyze the alignment of probe subspaces using an adapted version of SVCCA, which lets us test whether ground-truth-, output-related supervision gives rise to shared or distinct counting directions. We also perform causal steering intervention and prove that strengthening the count direction does improve model performance. Our contributions are as follows. \textbf{(1)} We construct four synthetic counting datasets paired with a real-world benchmark, enabling controlled evaluation across diverse counting regimes (Fig~\ref{fig:pipeline_labeled}). \textbf{(2)} Building on \citet{sun2025probarithlang}'s multi-probe framework for LLM arithmetic, we extend the paradigm to VLM counting, probing three supervision targets across both vision-encoder and text-decoder representations (Figs~\ref{fig:combined_probing_results};\ref{fig:detector_f1}). \textbf{(3)} We show via SVCCA that ground-truth and output signals occupy substantially misaligned readout subspaces, especially for nonlinear probes (Fig~\ref{fig:combined_wsvcca_results}), and confirm causal relevance via activation steering: strengthening the count direction improves accuracy while random directions degrade it. \textbf{(4)} We translate these insights into a detector-guided self-correction method that improves counting accuracy by up to $15.6$ absolute percentage points without fine-tuning (Table~\ref{tab:synthetic_self_correction}).

\section{Related Work}
\label{sec:related}

\paragraph{VLM counting and reliability.}
Counting has long been used as a stress test for visual reasoning, from VQA-style benchmarks such as TallyQA~\citep{acharya2019tallyqa} and everyday-scene counting~\citep{chattopadhyay2017counting, weng2025visnumbench} to recent CLIP-era analysis~\citep{paiss2023teaching}. Broader multimodal evaluations report systematic weaknesses in grounding and object fidelity, including poor counting performance~\citep{guo2025vlmcantcount, vo2025vlmbiased}, hallucination~\citep{li2023pope,fu2023mme}, and weak perception-reasoning alignment~\citep{liu2023mmbench,yu2023mmvet,yue2023mmmu, vo2025vlmbiased}. Our work is complementary: rather than proposing a new benchmark, we test whether hidden activations already contain recoverable signals that can identify and correct counting failures.

\paragraph{Probing and Inference-time correction.}
Probing methods are widely used to test which properties are encoded in activations~\citep{alain2016understanding,belinkov2022probing}. In NLP, probes have been used for syntax~\citep{hewitt2019structural}, pipeline-like linguistic structure~\citep{tenney2019bert}, and factual associations~\citep{meng2022rome}. Recent work also studies latent confidence and truth-related structure in activations~\citep{kadavath2022know,burns2023discovering}, as well as numeracy-specific representations~\citep{heinzerlingmonotonic,elshangitigeometry}. In VLMs, probes were used to both analyze the counting mechanism~\citep{hasani2025countingmechanism} and localize bottleneck components that limits counting performance~\citep{alghisi2025vlmcounting}. 
Inference-time intervention has shown that internal signals can improve factuality without parameter updates~\citep{li2023iti}. In parallel, test-time reasoning and correction methods such as self-consistency~\citep{wang2023selfconsistency}, Tree-of-Thought~\citep{yao2023tree}, ReAct~\citep{yao2023react}, Self-Refine~\citep{madaan2023self}, Reflexion~\citep{shinn2023reflexion}, and StateAct~\citep{rozanov-rei-2025-stateact} improve outputs through search or iterative feedback. Closest to our work is \cite{sun2025probarithlang}, which probes LLM hidden states for ground-truth, output, and binary-correctness signals during arithmetic reasoning and uses the resulting error detectors for selective re-prompting.


\section{Analysis: Can VLMs internally count objects?}
\label{sec:analysis}

Counting is an important instance of reasoning; it has been proven to be challenging for VLMs~\citep{guo2025vlmcantcount}. In this paper, we investigate the counting mechanism in VLMs. We formalize the counting task as an input image $I$ and a textual query $q$, the task requires the model $\mathcal{M}$ to produce a count estimate $\hat{y}$ such that $\hat{y} = \mathcal{M}(I, q)$, where $y$ denotes the ground-truth count and $y, \hat{y} \in \{1, 2, \ldots, 9\}$.

\subsection{Probe Architectures}
\label{sec:probes}

Let $x_l \in \mathbb{R}^{d}$ denote the final text-token representation at layer $l$, where $d$ is the hidden dimension of the VLM. A probe is a mapping from $x_l$ to a target label. We consider three probing tasks. First, a \textbf{ground-truth count probe} maps $x_l$ to the true count $y$. Second, an \textbf{output-count probe} maps the same representation $x_l$ to the model's own predicted count $\hat{y}$. These two probes have the same architecture but are trained with different supervision (we illustrate below what architectures investigated), allowing us to compare what aspects of the hidden state align with the correct answer versus the produced answer. Third, for error detection, we train a separate \textbf{error probe} to predict whether the model's answer is correct: 
$e = \mathbb{I}[\hat{y} \neq y] \in \{0,1\} $,
where $e=1$ indicates an incorrect counting prediction.
Thus, the count probes solve multi-class classification over count labels, while the error probe solves binary classification over correctness labels. 

We employ four probe types of increasing expressivity~\citep{sun2025probarithlang}.
The \textbf{circular probe} projects $\mathbf{x}_l$ onto a learned 2D plane
via $\mathbf{w}_1, \mathbf{w}_2 \in \mathbb{R}^{d}$,
encoding the digit in the angle:
$\hat{y} = \tfrac{10}{2\pi}\,\operatorname{atan2}(\mathbf{w}_1^\top \mathbf{x}_l,\, \mathbf{w}_2^\top \mathbf{x}_l)$.
The \textbf{linear probe} applies an affine map
$\hat{y} = \mathbf{w}^\top \mathbf{x}_l + b$,
trained with $\ell_2$ regularization and rounded at inference.
The \textbf{logistic regression probe} assigns per-class weights:
$\hat{y} = \arg\max_i\, (\mathbf{w}_i^\top \mathbf{x}_l)$.
The \textbf{MLP probe} adds a single hidden layer (ReLU, 512 units):
$\hat{y} = \arg\max_i\, (\mathbf{W}_2^\top \operatorname{ReLU}(\mathbf{W}_1^\top \mathbf{x}_l + \mathbf{b}_1) + \mathbf{b}_2)$.
All classification probes are trained with cross-entropy; the circular probe with smooth $\ell_1$ loss.

\subsection{Measuring Representational Alignment via SVCCA.}
\label{sec:svcca}

Probe accuracy tells us whether a representation contains information about a target. However, accuracy alone does not tell us whether two targets are decoded from the {same} internal directions. This distinction is central to our question. A layer may simultaneously contain information about the true count $y$ and the model output $\hat{y}$, yet the two may rely on different subspaces. In that case, the model may internally encode the correct quantity while organizing its final prediction around a different representational direction. Conversely, if probes trained on $y$ and $\hat{y}$ rely on highly similar subspaces, then the model's final answer is more closely aligned with the internal count information available at that layer.

To test this, we compare pairs of probes with the same architecture trained on the same hidden states but with different supervision: one probe predicts the ground-truth count $y$, and the other predicts the model output $\hat{y}$. We then ask whether the two probes read out their information from similar directions (subspaces) in representation space.

We quantify subspace alignment using Singular Vector Canonical Correlation Analysis (SVCCA;~\citep{raghu2017svcca}); full mathematical details are given in Appendix~\ref{Appendix: SVCCA details}. Given the weight matrices $W_{\mathrm{gt}}, W_{\mathrm{out}} \in \mathbb{R}^{c \times d}$ of two probes trained on ground-truth and model-output counts respectively, we extract the top-$k$ left singular vectors from each, yielding orthonormal bases $U_{\mathrm{gt}}^{(k)}, U_{\mathrm{out}}^{(k)} \in \mathbb{R}^{d \times k}$. We then compute the canonical correlations $\rho_1, \dots, \rho_k = \sigma_1, \dots, \sigma_k\left({U_{\mathrm{gt}}^{(k)}}^\top U_{\mathrm{out}}^{(k)}\right)$ and summarize alignment as: 
$    \mathrm{SVCCA}(W_{\mathrm{gt}}, W_{\mathrm{out}}) = \frac{1}{k}\sum_{i=1}^{k} \rho_i.
$

This score has a simple interpretation. Values near $1$ indicate that the ground-truth and output probes rely on nearly the same directions in the hidden representation, suggesting that the model's final count is aligned with the internally available count information at that layer. Values near $0$ indicate that the two probes depend on nearly orthogonal subspaces, suggesting a stronger mismatch between information aligned with the correct count and information aligned with the produced answer.

Our use of SVCCA is motivated by the need for a basis-invariant comparison. Directly comparing probe weight matrices is not reliable, because two probes can implement similar subspaces using different but equivalent bases for the same subspace. By comparing the subspaces spanned by the principal singular vectors instead of raw parameters, SVCCA provides a more robust measure of whether ground-truth-aligned and output-aligned counting information are geometrically similar inside the model. While SVCCA was originally introduced to compare activation matrices across layers or networks~\citep{raghu2017svcca}, applying it directly to activations is uninformative in our setting: both probes read from \emph{identical} hidden states, so activation-level comparison is trivially maximal. The supervision-specific information instead resides in the probes' readout matrices, whose rows live in the same $d$-dimensional coordinate system of the shared frozen representation and are therefore directly comparable. To our knowledge, we are the first to apply SVCCA to probe weight matrices as a tool for comparing supervision-specific readout subspaces.

This analysis gives us a mechanistic tool for locating where counting failures emerge. If a layer shows strong decodability of the true count but weak alignment with the output probe, this suggests that correct quantity information is present but not faithfully carried into the final answer. In contrast, if the ground-truth count is weakly decodable, the counting failure likely originates earlier in the visual or multimodal representation itself.

\subsection{Experimental Setup}
\label{sec:setup}

\paragraph{Datasets.}
\label{sec:datasets}
We carefully design and construct four synthetic datasets, each containing $1170$ images ($512\times512$, white background) with $\{1, \ldots, 9\}$ non-overlapping objects placed on a shuffled grid at a fixed size. The four datasets vary the visual diversity of objects along color and shape (e.g., same color/different shape vs.\ different colors/different shapes, details in ~\ref{appendix:synthetic_datasets_details} Table~\ref{tab:dataset-properties}). Samples are equally distributed across nine count classes with an $80/20$ stratified train/test split. We additionally evaluate on CountBench~\citep{zhang2024countbench}, removing images with more than nine objects and using a $50/50$ stratified split to match our experimental setup.

\paragraph{Models.}
We evaluate four publicly available VLMs: \emph{InternVL2-1B}, \emph{InternVL2-4B}~\citep{chen2024internvl}, \emph{Qwen3-VL-2B-Instruct}, and \emph{Qwen3-VL-8B-Instruct}~\citep{bai2025qwen3vl}. All models are run with greedy decoding. Models are from HuggingFace Transformers~\citep{wolf2020transformers}.
All models are evaluated under zero-shot settings. All probes are trained for $2000$ epochs using a weighted cross-entropy loss to mitigate class imbalance. The layer yielding the best performance will be used in the subsequent experiments of the corresponding probe type.
\paragraph{Training Probes.} For each (model, dataset, probe-type, task) combination, we train four probe architectures, sweep over all layers, and characterize each probe by the layer that achieves the highest held-out F1 score. We train each probe on three tasks: ground-truth count (gt\_probe), output-count (output\_probe), and error probe.  

This process is repeated for all probes, datasets, layers, all four models, and averaged over three random seeds.

\begin{figure}[t] 
    \centering
    \begin{subfigure}[b]{1\linewidth} 
        \centering
        \includegraphics[width=\linewidth]{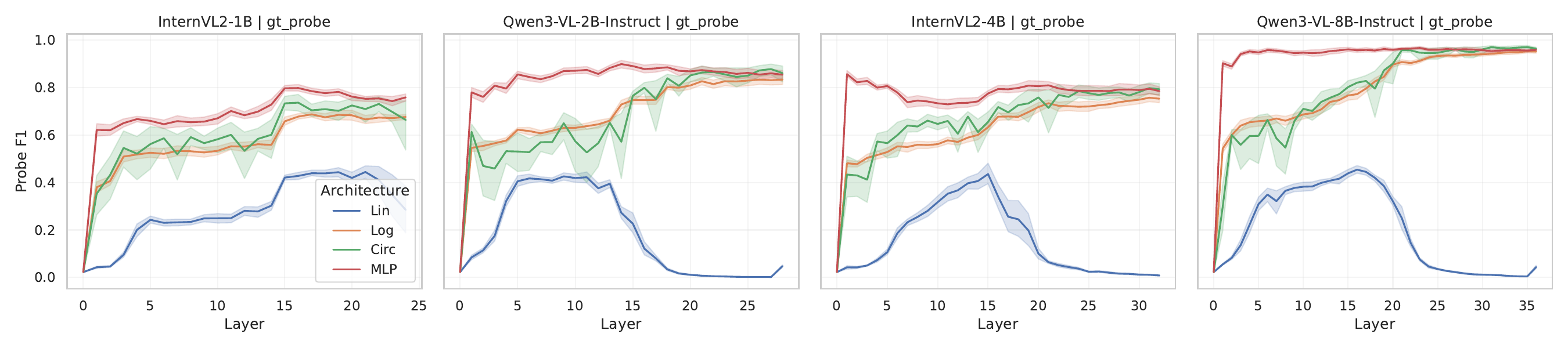}
        \caption{Layer-wise probing F1 score across model depths (ground truth objective).}
    \end{subfigure}
    
    \vspace{0.5cm} 

    \begin{subfigure}[b]{1\linewidth} 
        \centering
        \includegraphics[width=\linewidth]{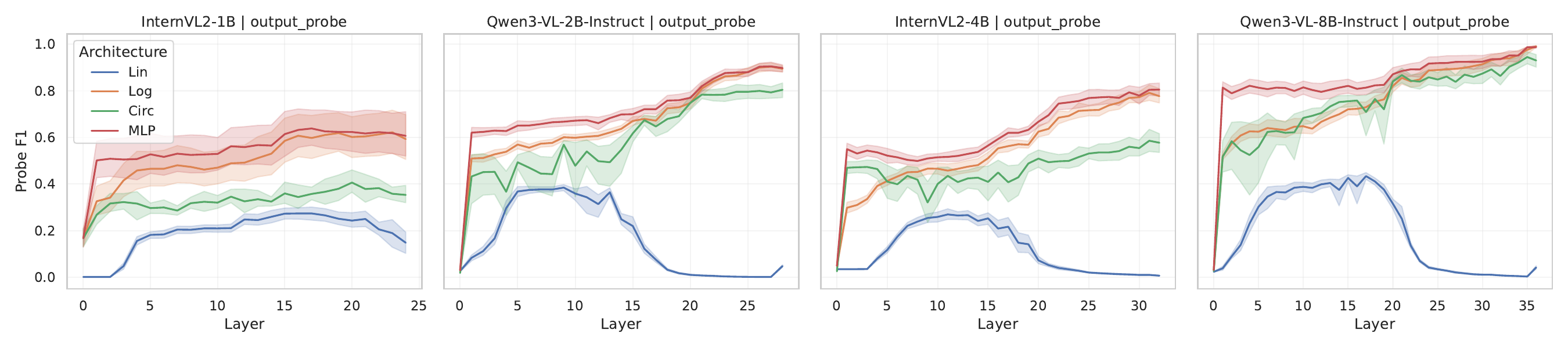}
        \caption{Layer-wise probing F1 score across model depths (model output objective).}
    \end{subfigure}

    \begin{subfigure}[b]{1\linewidth} 
        \centering
        \includegraphics[width=\linewidth]{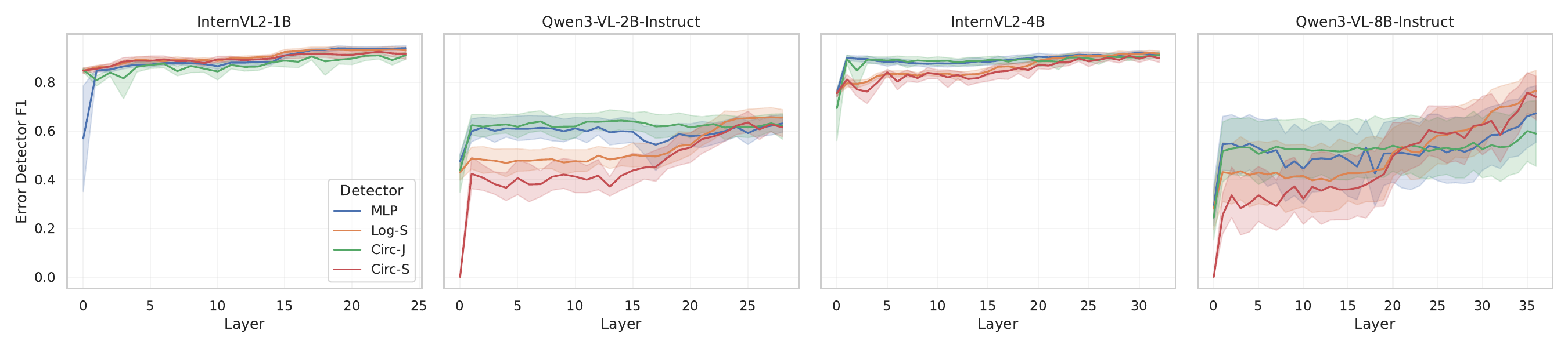}
    \caption{Error detector performance (F1-score) per layer.}
    \label{fig:detector_f1}
    \end{subfigure}

    \vspace{0.5cm} 
    
    \caption{Layer-wise probing and detector performance across model depth. Top(a): ground-truth probe F1 by layer. (b): output-supervised probe F1 by layer. Curves are aggregated across models, datasets, and the three random seeds. Bottom (c): Error detection }
    \label{fig:combined_probing_results}
\end{figure}

\subsection{Probing Results}
\label{sec: probing results}

Figure~\ref{fig:combined_probing_results} presents the results of training four probes on the activations of four models for two different objectives (ground truth and model output), aggregated across the four synthetic datasets, and three random seeds. Per-dataset results are provided in~\cref{fig:combined_probing_results__sing_col_sing_shape,fig:combined_probing_results__sing_col_diff_shape,fig:combined_probing_results__diff_col_sing_shape,fig:combined_probing_results__diff_col_diff_shape} in the Appendix \ref{Appendix: Detailed Results}.

 The single scalar affine probe is insufficient to decode count reliably, especially in later layers, while multiclass linear probes remain effective. All other architectures exceed 80\% F1, reaching near-perfect scores on Qwen3-VL-8B activations for both objectives. The fact that simpler probes such as Logistic and Circular achieve comparable performance to the MLP suggests that the recovered count information reflects genuine structure in the representations rather than an artifact of an expressive classifier learning the task independently~\citep{hewitt2019designing}. A notable temporal asymmetry emerges: ground-truth probes plateau at earlier layers, while output probes peak near the final layers, suggesting the model encodes the correct count well before it commits to a verbalized answer.

The exception is InternVL2-1B, where output probes remain flat around $60-65\%$, yet error detectors on the same activations achieve $>90\%$ F1 (Figure~\ref{fig:detector_f1}). This indicates that predicting \emph{whether} the model errs can be easier than predicting \emph{what} it will say; the binary error signal may occupy a more accessible subspace than the full output distribution.

For error detection, we train four detector types: MLP,
Logistic-Separately, Circular-Separately, and
Circular-Jointly~\citep{sun2025probarithlang}. The ``separately'' variants flag an error when two same-architecture probes (one trained on $y$, one on $\hat{y}$) disagree; the joint variant
trains a single model on the angular distance between two circular probes. Detectors on InternVL2 activations yield near-perfect F1, while those on Qwen3-VL activations perform lower, consistent with fewer training errors being available. As we show in~\S\ref{sec: self correction results} and Figure~\ref{fig:f1_vs_delta_scatter}, detector F1 is strongly correlated with downstream correction gain. Similar training and analysis have been done at the level of the last token and average tokens extracted from the vision encoder, more details in appendix ~\ref{appendix:vision_probing}. 

\begin{figure*}[t] 
    \centering
    
    \begin{subfigure}[b]{0.45\linewidth} 
        \centering
        \includegraphics[width=\linewidth]{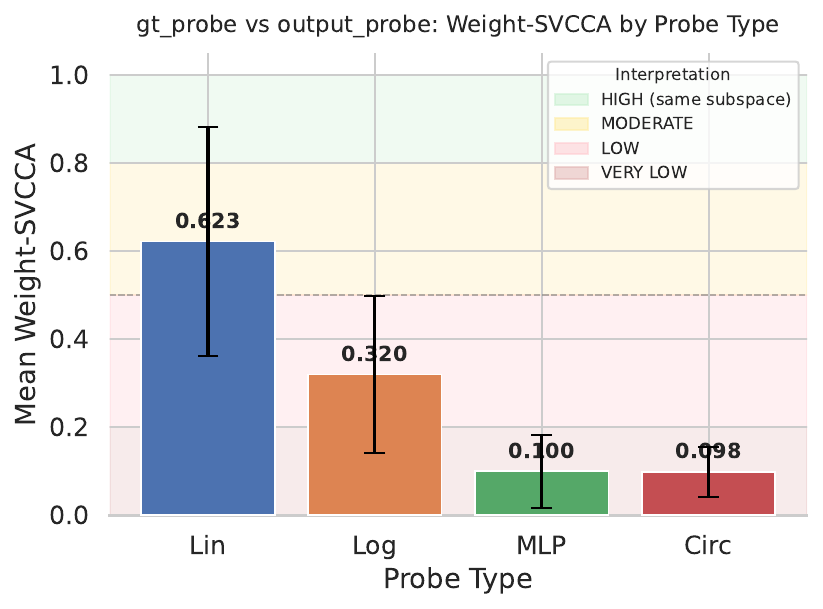}
        \caption{Mean weight-SVCCA by probe family.}
        \label{fig:svcca_by_probe_type_weight}
    \end{subfigure}
    \hfill 
    \begin{subfigure}[b]{0.45\linewidth} 
        \centering
        \includegraphics[width=\linewidth]{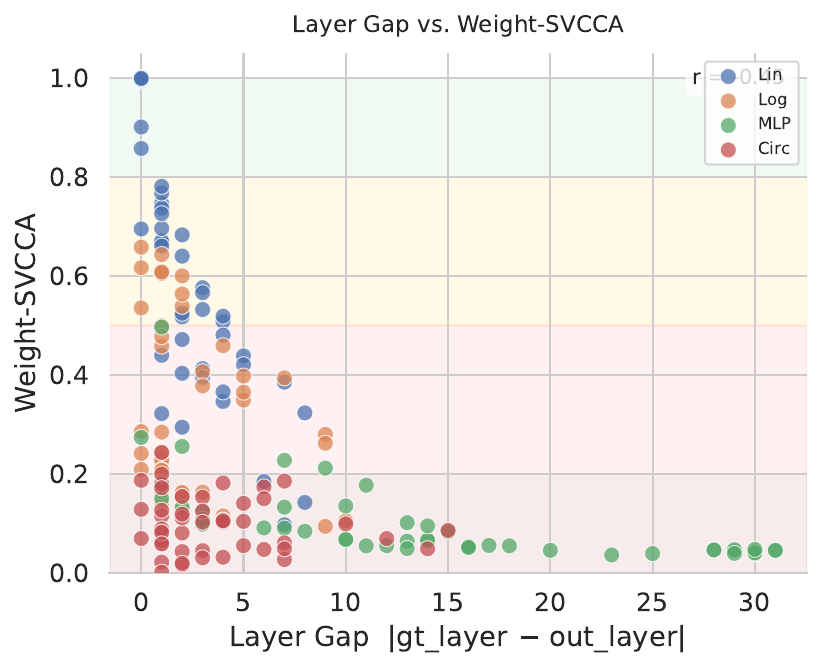}
        \caption{Weight-SVCCA versus best-layer gap.}
        \label{fig:svcca_layer_gap_scatter_weight}
    \end{subfigure}
    
    \caption{Weight-space SVCCA analysis of \texttt{gt\_probe} versus \texttt{output\_probe}. (\textbf{Left}) Mean SVCCA by probe family across the full 4-model, 4-dataset, 3-seed sweep. MLP and Circular probes lie in the near-orthogonal regime, while Linear probes show substantially higher alignment. (\textbf{Right}) SVCCA versus best-layer gap $|l_{gt}-l_{out}|$. Larger layer disagreement tends to coincide with lower alignment, but low SVCCA persists even at small layer gaps, indicating that subspace divergence is not reducible to depth mismatch alone.}
    \label{fig:combined_wsvcca_results}
\end{figure*}

\subsection{SVCCA: Subspace Divergence Between Ground-Truth and Output Probes.}
\label{sec:svcca-results}
SVCCA measures similarity between learned representations. In Figure~\ref{fig:combined_wsvcca_results}, we report SVCCA scores computed between the weights (Figure~\ref{fig:svcca_by_probe_type_activation} for activations) of the same probe family trained under two objectives: ground-truth count prediction and model-output count prediction. In Figure~\ref{fig:combined_wsvcca_results}(a), nonlinear probes (Circular and MLP) lie in a near-orthogonal regime (low SVCCA). Linear probes show higher alignment than nonlinear probes, but alignment remains limited, especially for Logistic regression. These findings support a subspace-misalignment hypothesis: count-relevant information is internally encoded, but only weakly coupled to the representation subspace that drives final verbalized outputs. To test whether this pattern is merely a layer-selection artifact, Figure~\ref{fig:combined_wsvcca_results}(b) plots SVCCA against the best-layer gap. Low SVCCA persists even at small layer gaps, indicating that subspace divergence is not reducible to depth mismatch alone, as discussed in prior findings~\citep{burns2023discovering, gekhman2025factualknowledge}

\section{Causal Steering Intervention}
\label{sec:causal-steering}
\begin{figure}[t]
    \centering
    \includegraphics[width=\linewidth]{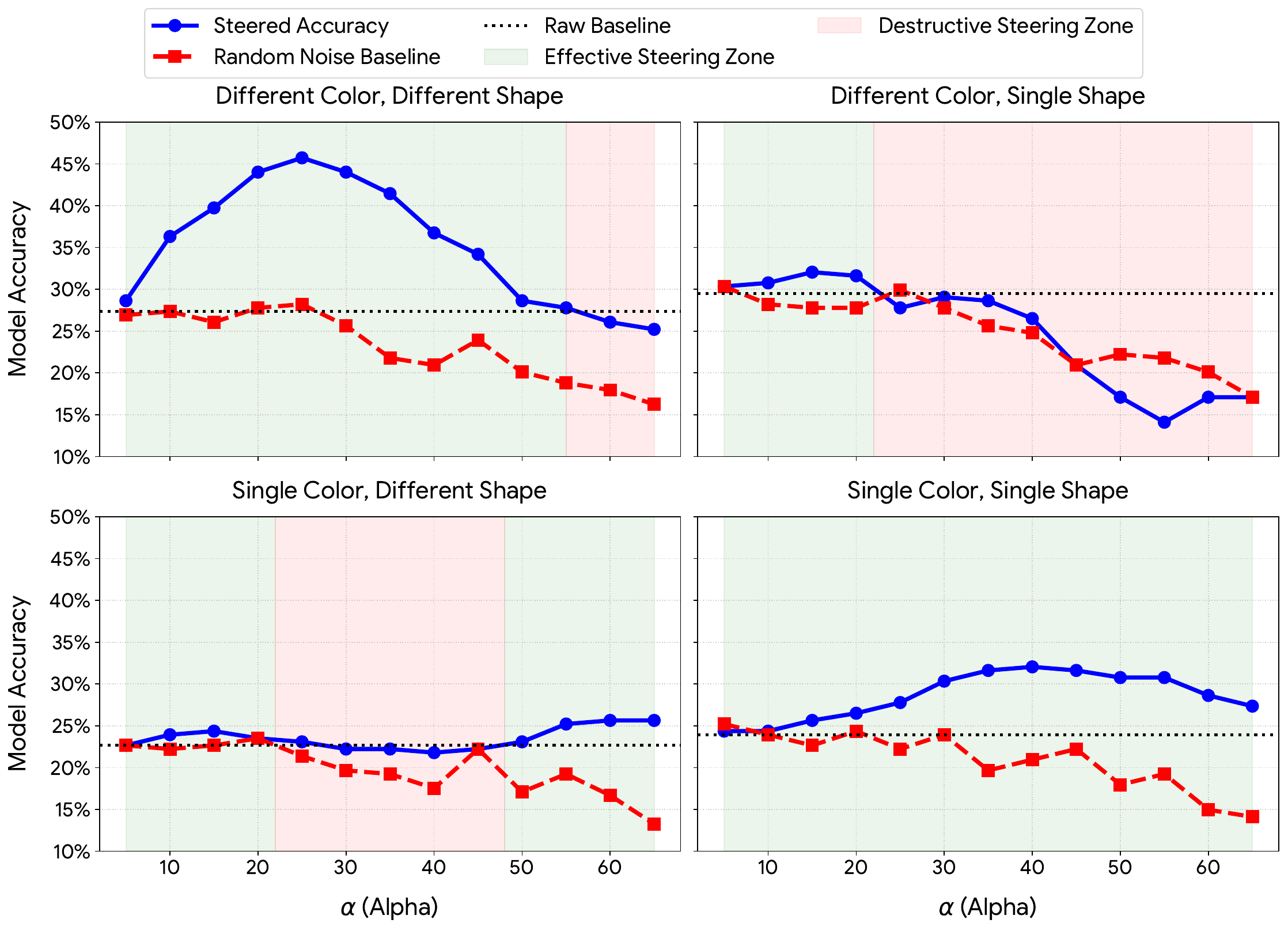}
    \caption{Causal steering results for \emph{InternVL2-1B} on four synthetic counting datasets. Green shading marks the \emph{effective steering zone} where probe-derived steering exceeds the baseline; pink shading marks the \emph{destructive zone} where steering falls below the baseline.}
    \label{fig:causal-steering-results}
\end{figure}

The probing and SVCCA analyses above establish that VLMs internally encode count information that diverges from their verbalized outputs. These results are correlational: the probe may recover a direction that is \emph{predictive} of count without being \emph{causally} involved in generating it. To distinguish these possibilities, we perform a direct activation-steering experiment on the full synthetic test split of all four synthetic datasets.
\paragraph{Method.}
For each sample we first run the unmodified model to obtain a baseline prediction $\hat{y}$ and record the baseline accuracy (Figure~\ref{fig:causal-steering-results}). We then rerun the same sample with a single forward-hook intervention applied at the best layer $l^*$ of the saved Logistic \texttt{gt\_probe}. Let $x_{l^*}\!\in\!\mathbb{R}^{d}$ denote the final prompt-token hidden state at layer $l^*$, and let $w_c$ be the probe weight vector for class $c$. We define the steering direction as the normalized pairwise difference between the ground-truth and baseline class directions:
\begin{equation}
d =
\frac{w_y - w_{\hat{y}}}{\left\| w_y - w_{\hat{y}} \right\|_2},
\qquad
\tilde{x}_{l^*} = x_{l^*} + \alpha \cdot \mathrm{RMS}(x_{l^*}) \cdot d,
\qquad
\end{equation}
$\mathrm{RMS}(x) = \frac{\left\| x \right\|_2}{\sqrt{d}},$

where $\alpha$ controls steering strength and RMS-scaling ensures the perturbation is proportional to the local activation norm. As a control, we run the same experiment replacing $d$ with a randomly sampled unit vector of the same dimensionality. Each $\alpha$ is swept over a range from $5$ to $65$.
\paragraph{Results.}
Figure~\ref{fig:causal-steering-results} plots model accuracy as a function of $\alpha$ for both the probe-derived (blue) and random-direction (red) interventions. Three observations emerge:
\textbf{(1) Probe-derived steering causally improves counting.}
On all four datasets, steering along the probe direction lifts accuracy above the unsteered baseline within a broad effective range (green region). The clearest gain appears on \texttt{diff\_col\_diff\_shape}, where accuracy rises from a baseline of ${\sim}27\%$ to a peak near $46\%$ at moderate $\alpha$, corresponding to an absolute gain of roughly $19$ percentage points. On \texttt{sing\_col\_sing\_shape} and \texttt{diff\_col\_sing\_shape}, more modest but consistent gains (${\sim}6$--$8$ pp) persist across a range of moderate intervention strengths. These results demonstrate that the direction recovered by the probe is not merely predictive of count but plays a causal role in the model's generation.
\textbf{(2) Random steering monotonically degrades performance.}
Across all four datasets, the random-direction control either stays near the baseline at small $\alpha$ or steadily degrades as $\alpha$ grows (red curves entering the pink region), confirming that the improvements in (1) are specific to the probe-derived direction rather than an artifact of generic activation perturbation. The widening gap between the blue and red curves as $\alpha$ increases further corroborates the directional specificity of the causal effect.
\textbf{(3) Excessive steering is destructive.}
On all datasets, there exists a critical $\alpha$ beyond which even probe-derived steering begins to hurt. On \texttt{sing\_col\_diff\_shape}, the blue curve enters the destructive zone around $\alpha\!=\!45$, and on \texttt{diff\_col\_sing\_shape}, gains plateau early. This non-monotonicity is consistent with a local-linear interpretation: the probe direction is a good first-order approximation of the count subspace, but large perturbations push the hidden state beyond the regime where this linear approximation holds.

\paragraph{Implications.}
The causal steering experiment indicates that probe-derived count directions can influence
generation, but it also reveals that direct activation intervention depends on two factors: dataset geometry and steering magnitude $\alpha$. This motivates a less invasive use of the same representational insight: rather
than modifying hidden states directly, we use the activation-level error signal to decide
when the model should reconsider its answer. Figure~\ref{fig:f1_vs_delta_scatter} supports this choice by showing a
strong positive correlation between detector F1 and correction gain, indicating that reliable
internal failure prediction translates into effective downstream correction. We therefore turn
to a detector-guided self-correction method that treats the probe as an intervention gate
rather than as a steering mechanism. Next, we introduce detector-guided self-correction in
Section~\ref{sec:self correction}

\section{Detector-Guided Self-Correction Method}
\label{sec:self correction}

Let $x _{l^*}\in \mathbb{R}^d$ be the activation extracted from last non-padding token from layer $l^*$ of VLM $\mathcal{M}$ for image-query pair $(I, q)$, where $l^* = \arg\max_{l}\,\mathrm{F1}(D^{(l)}, \mathcal{D}_{\mathrm{val}})$ is the layer yielding the best detector performance on a held-out validation set $\mathcal{D}_{\mathrm{val}}$. Let $\hat{y}^{(1)} = \mathcal{M}(I, q)$ be the first-pass prediction. A detector $D^{(l^*)} : \mathbb{R}^d \to [0,1]$ estimates the error probability: \\

\begin{equation}
s = D^{(l^*)}(x_{l^*}) \approx \mathbb{P}\!\left(\hat{y}^{(1)} \neq y \mid x_{l^*}\right)
\end{equation}

and triggers a correction when $s$ exceeds a threshold $\tau$\footnote{here, $\tau$ is fixed to a constant equal $0.5$}:
\begin{equation}
    \hat{y} = \begin{cases} \mathcal{M}\!\left(I,\; f_{\mathrm{corr}}(q, \hat{y}^{(1)})\right) ;& \text{if } $s$ \ge \tau \\ \hat{y}^{(1)} ; & \text{otherwise} \end{cases}
\end{equation} 
where $f_{\mathrm{corr}}(q, \hat{y}^{(1)})$ is a correction prompt that supplies the original response and asks the model to reconsider its answer. The correction prompt is in the Appendix~\ref{app:correction_prompt}. 
\\

\paragraph{Baselines.} We compare against two baselines. \emph{Always} \emph{Reprompt} applies $f_{\mathrm{corr}}$ to all samples unconditionally. This is the maximum budget baseline. \emph{Random-$K$} matches the detector's intervention budget $K = |\{i: D(x_{l^*,i}) \ge \tau\}|$ but selects samples at random, isolating the contributions of learned selectivity over random reprompting. An additional baseline (Entropy-guided) is reported in Table~\ref{appendix:entropy_vs_probe}.

\subsection{Proposed Method Results}
\label{sec: self correction results}

\begin{table}[t]
\centering
\scriptsize
\setlength{\tabcolsep}{2pt} 
\resizebox{\columnwidth}{!}{
\begin{tabular}{lccccc}
\toprule
\textbf{Dataset} & \textbf{Model} & \textbf{Raw} & \textbf{Always Repr.} & \textbf{Random-K} & \textbf{Ours ($\Delta$)} \\
\midrule
\multirow{4}{*}{\texttt{DC-DS}}
& IVL2-1B  & 26.50 & \textcolor{negred}{23.50} & \textcolor{negred}{23.97} & \textbf{31.80} (\textcolor{posgreen}{+5.31}) \\
& IVL2-4B  & 34.62 & 50.85 & 45.73 & \textbf{50.21} (\textcolor{posgreen}{+15.60}) \\
& Q3VL-2B  & 71.51 & 71.51 & 71.69 & \textbf{72.04} (\textcolor{posgreen}{+0.53}) \\
& Q3VL-8B  & 79.34 & 84.33 & 80.31 & \textbf{83.76} (\textcolor{posgreen}{+4.42}) \\
\midrule
\multirow{4}{*}{\texttt{DC-SS}}
& IVL2-1B  & 29.77 & \textcolor{negred}{28.21} & \textcolor{negred}{28.45} & \textbf{35.43} (\textcolor{posgreen}{+5.66}) \\
& IVL2-4B  & 39.89 & 44.16 & 43.02 & \textbf{43.95} (\textcolor{posgreen}{+4.06}) \\
& Q3VL-2B  & 71.94 & 72.93 & 72.65 & \textbf{72.83} (\textcolor{posgreen}{+0.89}) \\
& Q3VL-8B  & 92.31 & 94.44 & 92.59 & \textbf{94.41} (\textcolor{posgreen}{+2.10}) \\
\midrule
\multirow{4}{*}{\texttt{SC-DS}}
& IVL2-1B  & 24.22 & 24.22 & \textcolor{negred}{23.50} & \textbf{30.88} (\textcolor{posgreen}{+6.66}) \\
& IVL2-4B  & 43.02 & 47.58 & 46.79 & \textbf{50.61} (\textcolor{posgreen}{+7.59}) \\
& Q3VL-2B  & 61.40 & 62.11 & \textcolor{negred}{61.18} & \textbf{62.46} (\textcolor{posgreen}{+1.07}) \\
& Q3VL-8B  & 66.95 & 77.49 & 71.94 & \textbf{77.49} (\textcolor{posgreen}{+10.54}) \\
\midrule
\multirow{4}{*}{\texttt{SC-SS}}
& IVL2-1B  & 23.65 & 30.48 & 28.77 & \textbf{35.11} (\textcolor{posgreen}{+11.47}) \\
& IVL2-4B  & 39.74 & 44.73 & 43.20 & \textbf{45.37} (\textcolor{posgreen}{+5.63}) \\
& Q3VL-2B  & 69.80 & 70.94 & 70.48 & \textbf{71.62} (\textcolor{posgreen}{+1.82}) \\
& Q3VL-8B  & 90.60 & 92.17 & 90.92 & \textbf{92.59} (\textcolor{posgreen}{+1.99}) \\
\midrule
\multirow{4}{*}{\texttt{CountBench}}
& IVL2-1B  & 40.15 & 49.24 & 46.78 & \textbf{46.97} (\textcolor{posgreen}{+6.82}) \\
& IVL2-4B  & 51.33 & 54.73 & 51.66 & \textbf{55.11} (\textcolor{posgreen}{+3.79}) \\
& Q3VL-2B  & 72.92 & 78.79 & 73.67 & \textbf{79.31} (\textcolor{posgreen}{+6.39}) \\
& Q3VL-8B  & 79.92 & 89.02 & 81.11 & \textbf{88.59} (\textcolor{posgreen}{+8.66}) \\
\bottomrule
\end{tabular}%
}
\caption{Mean counting accuracy (\%) aggregated over 3 seeds and 4 detectors. \textbf{Always Repr.} is naive always-re-prompt; it does apply correction (reprompting) to $100\%$ of the evaluated samples. \textbf{Random-K} is budget-matched random intervention; \textbf{Ours} is detector-guided correction. $\Delta$ reports only \textbf{Ours} $-$ \textbf{Before} (pp). Values in \textcolor{negred}{red} indicate performance degradation compared to \textbf{Raw}.}
\label{tab:synthetic_self_correction}
\end{table}

Table~\ref{tab:synthetic_self_correction} reports the performance of the zero-shot raw model, the \emph{Always Reprompt} baseline, the \emph{Random-K} baseline, and our method across four synthetic datasets and four VLMs. In the \emph{Always Reprompt} setting, every sample is reprompted unconditionally (i.e, $100\%$ of the samples), whereas \emph{Random-K} reprompts the same number $K$ of samples as our probe flags, but selects them uniformly at random.

 
 We observe that, for almost all models, the dataset with a single color and varying shapes is the most difficult. The only exception is the InternVL2-4B model. 

While \emph{Always Reprompt} outperforms \emph{Ours} in several configurations, it is not a practical strategy for two reasons. First, it is computationally expensive, as it unconditionally requires a second forward pass on the entire dataset. Second, indiscriminate reprompting is unreliable: it can flip previously correct answers, degrading overall performance (Table~\ref{tab:synthetic_self_correction}). In contrast, \emph{Ours} recovers most of the gains achieved by \emph{Always Reprompt} while selectively reprompting only a small subset of samples, avoiding both the computational overhead and the risk of degradation. It consistently surpasses the \emph{Random-K} baseline. Overall, our approach yields improvements of up to $15.6\%$ (avg $5.3\%$). The largest gains occur for the InternVL2 family, while the smallest gains are observed for Qwen3-VL-2B; these trends align closely with the quality of the error-detection probe (Figure~\ref{fig:detector_f1}). Figure~\ref{fig:f1_vs_delta_scatter} further illustrates how probe performance on error detection correlates with self-correction gains, showing that the largest improvements are obtained when the probe's F1-score exceeds $90\%$ (with strong correlation $\rho = 0.803$). Table~\ref{tab:detectors-stratified} provides a summary of correction gains across detector types. We can see several patterns. First, no single detector consistently dominates, suggesting that detector choice is not critical and the method is robust to error detector architecture. Second, the discrepancy between VLM families is larger than it between detectors: InternVL2 benefits substantially (up to $+15.95$ pp) while Qwen3-VL-2B gains is ($+0.43$ to $+2.42$ pp) independent of probe choice, suggesting that the bottleneck for correction is the quality of the error signal in the activation rather than the detector architecture itself. Table A.~\ref{tab:full_detector_analysis} reports true-positive correction rates and false positive preservation rates for each detector. Additional prompts are reported in~\ref{appendix:prompt_comparison_internvl1b}. By repeating the full pipeline on the CountBench dataset, we prove that our findings hold beyond synthetic datasets.

\section{Conclusion}
\label{sec:conclusion}
In this paper, we introduce the first study that examines how simple probes can predict the ground-truth count, the model-predicted count, and the model likelihood of error from VLMs' internal representations. We further demonstrate that these simple probes can accurately anticipate when a VLM is about to generate an incorrect count. A complementary SVCCA analysis shows that probes trained on ground-truth counts and those trained on model outputs occupy  misaligned readout subspaces, indicating a divergence between true and output count representations during generation. We further validate our findings using a causal steering intervention, proving that strengthening the count-identified probes' direction does improve model counting performance. Finally, we integrate these insights into a method that uses simple probes for error detection and reprompting-based correction, achieving up to a $15.6\%$ improvement in counting performance.

\section{Limitations and Future Work}
\label{sec:limits}
The limitations are that we study only models up to 8B parameters due to computational constraints, use synthetic data whose transfer to real-world settings is still unverified, cover only two VLM families, conduct a causal intervention experiment with only one model, and restrict counting tasks to the $1$–$9$ range. Future work should evaluate larger models, real-world generalization, additional VLM families, and harder counting regimes beyond $10$ objects.

\bibliography{custom}

@inproceedings{acharya2019tallyqa,
  title     = {{TallyQA}: Answering Complex Counting Questions},
  author    = {Acharya, Manoj and Kafle, Kushal and Kanan, Christopher},
  booktitle = {Proceedings of the AAAI Conference on Artificial Intelligence},
  volume    = {33},
  pages     = {8076--8084},
  year      = {2019},
  doi       = {10.1609/aaai.v33i01.33018076}
}

@inproceedings{chattopadhyay2017counting,
  title     = {Counting Everyday Objects in Everyday Scenes},
  author    = {Chattopadhyay, Prithvijit and Vedantam, Ramakrishna and
               Selvaraju, Ramprasaath R. and Batra, Dhruv and Parikh, Devi},
  booktitle = {Proceedings of the IEEE Conference on Computer Vision and
               Pattern Recognition (CVPR)},
  pages     = {1135--1144},
  year      = {2017}
}

@inproceedings{paiss2023teaching,
  title     = {Teaching {CLIP} to Count to Ten},
  author    = {Paiss, Roni and Ephrat, Ariel and Tov, Omer and Zada, Shiran and
               Mosseri, Inbar and Irani, Michal and Dekel, Tali},
  booktitle = {Proceedings of the IEEE/CVF International Conference on
               Computer Vision (ICCV)},
  pages     = {3323--3333},
  year      = {2023}
}

@misc{zhang2024countbench,
      title={Teaching CLIP to Count to Ten}, 
      author={Roni Paiss and Ariel Ephrat and Omer Tov and Shiran Zada and Inbar Mosseri and Michal Irani and Tali Dekel},
      year={2023},
      eprint={2302.12066},
      archivePrefix={arXiv},
      primaryClass={cs.CV},
      url={https://arxiv.org/abs/2302.12066}, 
}

@article{belinkov2022probing,
  title     = {Probing Classifiers: Promises, Shortcomings, and Advances},
  author    = {Belinkov, Yonatan},
  journal   = {Computational Linguistics},
  volume    = {48},
  number    = {1},
  pages     = {207--219},
  year      = {2022},
  publisher = {MIT Press},
  doi       = {10.1162/coli_a_00422}
}

@inproceedings{tenney2019bert,
  title     = {{BERT} Rediscovers the Classical {NLP} Pipeline},
  author    = {Tenney, Ian and Das, Dipanjan and Pavlick, Ellie},
  booktitle = {Proceedings of the 57th Annual Meeting of the Association for
               Computational Linguistics},
  pages     = {4593--4601},
  year      = {2019},
  address   = {Florence, Italy},
  publisher = {Association for Computational Linguistics},
  doi       = {10.18653/v1/P19-1452}
}

@inproceedings{hewitt2019structural,
  title     = {A Structural Probe for Finding Syntax in Word Representations},
  author    = {Hewitt, John and Manning, Christopher D.},
  booktitle = {Proceedings of the 2019 Conference of the North {A}merican
               Chapter of the Association for Computational Linguistics:
               Human Language Technologies, Volume 1 (Long and Short Papers)},
  pages     = {4129--4138},
  month     = jun,
  year      = {2019},
  address   = {Minneapolis, Minnesota},
  publisher = {Association for Computational Linguistics},
  doi       = {10.18653/v1/N19-1419}
}

@inproceedings{meng2022rome,
  title     = {Locating and Editing Factual Associations in {GPT}},
  author    = {Meng, Kevin and Bau, David and Andonian, Alex and Belinkov, Yonatan},
  booktitle = {Advances in Neural Information Processing Systems},
  volume    = {35},
  pages     = {17359--17372},
  year      = {2022},
  publisher = {Curran Associates, Inc.}
}

@inproceedings{raghu2017svcca,
  title     = {{SVCCA}: Singular Vector Canonical Correlation Analysis for
               Deep Learning Dynamics and Interpretability},
  author    = {Raghu, Maithra and Gilmer, Justin and Yosinski, Jason and
               Sohl-Dickstein, Jascha},
  booktitle = {Advances in Neural Information Processing Systems},
  volume    = {30},
  year      = {2017},
  publisher = {Curran Associates, Inc.}
}

@inproceedings{li2023iti,
  title     = {Inference-Time Intervention: Eliciting Truthful Answers from
               a Language Model},
  author    = {Li, Kenneth and Patel, Oam and Vi{\'e}gas, Fernanda and
               Pfister, Hanspeter and Wattenberg, Martin},
  booktitle = {Advances in Neural Information Processing Systems},
  volume    = {36},
  year      = {2023},
  publisher = {Curran Associates, Inc.}
}

@inproceedings{madaan2023self,
  title     = {Self-Refine: Iterative Refinement with Self-Feedback},
  author    = {Madaan, Aman and Tandon, Niket and Gupta, Prakhar and
               Hallinan, Skyler and Gao, Luyu and Wiegreffe, Sarah and
               Alon, Uri and Dziri, Nouha and Prabhumoye, Shrimai and
               Yang, Yiming and Gupta, Shashank and Welleck, Sean and
               Yazdanbakhsh, Amir and Clark, Peter},
  booktitle = {Advances in Neural Information Processing Systems},
  volume    = {36},
  year      = {2023},
  publisher = {Curran Associates, Inc.}
}

@inproceedings{shinn2023reflexion,
  title     = {Reflexion: Language Agents with Verbal Reinforcement Learning},
  author    = {Shinn, Noah and Cassano, Federico and Gopinath, Ashwin and
               Narasimhan, Karthik and Yao, Shunyu},
  booktitle = {Advances in Neural Information Processing Systems},
  volume    = {36},
  year      = {2023},
  publisher = {Curran Associates, Inc.}
}

@misc{sun2025probarithlang,
      title={Probing for Arithmetic Errors in Language Models}, 
      author={Yucheng Sun and Alessandro Stolfo and Mrinmaya Sachan},
      year={2025},
      eprint={2507.12379},
      archivePrefix={arXiv},
      primaryClass={cs.CL},
      url={https://arxiv.org/abs/2507.12379}, 
}

@inproceedings{elshangitigeometry,
    title = "The Geometry of Numerical Reasoning: Language Models Compare Numeric Properties in Linear Subspaces",
    author = "El-Shangiti, Ahmed Oumar  and
      Hiraoka, Tatsuya  and
      AlQuabeh, Hilal  and
      Heinzerling, Benjamin  and
      Inui, Kentaro",
    editor = "Chiruzzo, Luis  and
      Ritter, Alan  and
      Wang, Lu",
    booktitle = "Proceedings of the 2025 Conference of the Nations of the Americas Chapter of the Association for Computational Linguistics: Human Language Technologies (Volume 2: Short Papers)",
    month = apr,
    year = "2025",
    address = "Albuquerque, New Mexico",
    publisher = "Association for Computational Linguistics",
    url = "https://aclanthology.org/2025.naacl-short.47/",
    doi = "10.18653/v1/2025.naacl-short.47",
    pages = "550--561",
    ISBN = "979-8-89176-190-2"
}

@inproceedings{heinzerlingmonotonic,
    title = "Monotonic Representation of Numeric Attributes in Language Models",
    author = "Heinzerling, Benjamin  and
      Inui, Kentaro",
    editor = "Ku, Lun-Wei  and
      Martins, Andre  and
      Srikumar, Vivek",
    booktitle = "Proceedings of the 62nd Annual Meeting of the Association for Computational Linguistics (Volume 2: Short Papers)",
    month = aug,
    year = "2024",
    address = "Bangkok, Thailand",
    publisher = "Association for Computational Linguistics",
    url = "https://aclanthology.org/2024.acl-short.18/",
    doi = "10.18653/v1/2024.acl-short.18",
    pages = "175--195"
}

@misc{guo2025vlmcantcount,
      title={Your Vision-Language Model Can't Even Count to 20: Exposing the Failures of VLMs in Compositional Counting}, 
      author={Xuyang Guo and Zekai Huang and Zhenmei Shi and Zhao Song and Jiahao Zhang},
      year={2025},
      eprint={2510.04401},
      archivePrefix={arXiv},
      primaryClass={cs.CV},
      url={https://arxiv.org/abs/2510.04401}, 
}

@inproceedings{alain2016understanding,
  title     = {Understanding Intermediate Layers Using Linear Classifier Probes},
  author    = {Alain, Guillaume and Bengio, Yoshua},
  booktitle = {International Conference on Learning Representations (ICLR) Workshop},
  year      = {2017},
  note      = {arXiv:1610.01644}
}

@inproceedings{li2023blip2,
  title     = {BLIP-2: Bootstrapping Language-Image Pre-training with Frozen Image Encoders and Large Language Models},
  author    = {Li, Junnan and Li, Dongxu and Savarese, Silvio and Hoi, Steven C. H.},
  booktitle = {Proceedings of the 40th International Conference on Machine Learning},
  series    = {Proceedings of Machine Learning Research},
  volume    = {202},
  pages     = {19730--19742},
  year      = {2023},
  publisher = {PMLR}
}

@article{li2023pope,
  title   = {Evaluating Object Hallucination in Large Vision-Language Models: The {POPE} Benchmark},
  author  = {Li, Yifan and Du, Yifan and Zhou, Kun and Wang, Jinpeng and Zhao, Wayne Xin and others},
  journal = {arXiv preprint arXiv:2305.10355},
  year    = {2023}
}

@article{fu2023mme,
  title   = {{MME}: A Comprehensive Evaluation Benchmark for Multimodal Large Language Models},
  author  = {Fu, Chaoyou and Dai, Yujie and Luo, Yinpeng and Li, Liang and Ren, Shuhuai and Zhang, Runpeng and Wang, Zihan and Zhou, Chenyang and Shen, Yadong and Zhang, Meng and others},
  journal = {arXiv preprint arXiv:2306.13394},
  year    = {2023}
}

@article{liu2023mmbench,
  title   = {{MMBench}: Is Your Multi-modal Model an All-around Player?},
  author  = {Liu, Yang and Du, Wenhai and Zhang, Kai and Li, Xiaoxin and Hu, Jinghao and Liu, Xiang and Zhou, Zicheng and He, Yuan and Qiu, Zejun and others},
  journal = {arXiv preprint arXiv:2307.06281},
  year    = {2023}
}

@article{yu2023mmvet,
  title   = {{MM-Vet}: Evaluating Large Multimodal Models for Integrated Capabilities},
  author  = {Yu, Wenhao and Chen, Yiming and Wu, Xiang and He, Zhen and Liu, Yuhang and Zhao, Xu and others},
  journal = {arXiv preprint arXiv:2308.02490},
  year    = {2023}
}

@article{yue2023mmmu,
  title   = {{MMMU}: A Massive Multi-discipline Multimodal Understanding and Reasoning Benchmark for Expert AGI},
  author  = {Yue, Xi and Ni, Yuan and Zhang, Kai and Zheng, Tong and Liu, Yang and Gao, Wenxuan and others},
  journal = {arXiv preprint arXiv:2311.16502},
  year    = {2023}
}

@article{kadavath2022know,
  title   = {Language Models (Mostly) Know What They Know},
  author  = {Kadavath, Saurav and Schaeffer, Rylan and Kwon, Juhyeon and Mills, Katie and Yao, Alyssa and others},
  journal = {arXiv preprint arXiv:2207.05221},
  year    = {2022}
}

@article{burns2023discovering,
  title   = {Discovering Latent Knowledge in Language Models Without Supervision},
  author  = {Burns, Collin and Ye, Haotian and Klein, Dan and Steinhardt, Jacob},
  journal = {arXiv preprint arXiv:2212.03827},
  year    = {2023}
}

@inproceedings{wang2023selfconsistency,
  title     = {Self-Consistency Improves Chain of Thought Reasoning in Language Models},
  author    = {Wang, Xuezhi and Wei, Jason and Schuurmans, Dale and Le, Quoc and Chi, Ed and Narang, Sharan and Chowdhery, Aakanksha and Zhou, Denny},
  booktitle = {International Conference on Learning Representations},
  year      = {2023}
}

@article{yao2023tree,
  title   = {Tree of Thoughts: Deliberate Problem Solving with Large Language Models},
  author  = {Yao, Shunyu and Yu, Dian and Zhao, Jeffrey and Shafran, Izhak and Narasimhan, Karthik and Cao, Yuan and others},
  journal = {arXiv preprint arXiv:2305.10601},
  year    = {2023}
}

@inproceedings{yao2023react,
  title     = {{ReAct}: Synergizing Reasoning and Acting in Language Models},
  author    = {Yao, Shunyu and Zhao, Jeffrey and Yu, Dian and Du, Nan and Shafran, Izhak and Narasimhan, Karthik and Cao, Yuan},
  booktitle = {International Conference on Learning Representations},
  year      = {2023}
}

@inproceedings{chen2024internvl,
  title={Internvl: Scaling up vision foundation models and aligning for generic visual-linguistic tasks},
  author={Chen, Zhe and Wu, Jiannan and Wang, Wenhai and Su, Weijie and Chen, Guo and Xing, Sen and Zhong, Muyan and Zhang, Qinglong and Zhu, Xizhou and Lu, Lewei and others},
  booktitle={Proceedings of the IEEE/CVF Conference on Computer Vision and Pattern Recognition},
  pages={24185--24198},
  year={2024}
}

@misc{bai2025qwen3vl,
      title={Qwen3-VL Technical Report}, 
      author={Shuai Bai and Yuxuan Cai and Ruizhe Chen and Keqin Chen and Xionghui Chen and Zesen Cheng and Lianghao Deng and Wei Ding and Chang Gao and Chunjiang Ge and Wenbin Ge and Zhifang Guo and Qidong Huang and Jie Huang and Fei Huang and Binyuan Hui and Shutong Jiang and Zhaohai Li and Mingsheng Li and Mei Li and Kaixin Li and Zicheng Lin and Junyang Lin and Xuejing Liu and Jiawei Liu and Chenglong Liu and Yang Liu and Dayiheng Liu and Shixuan Liu and Dunjie Lu and Ruilin Luo and Chenxu Lv and Rui Men and Lingchen Meng and Xuancheng Ren and Xingzhang Ren and Sibo Song and Yuchong Sun and Jun Tang and Jianhong Tu and Jianqiang Wan and Peng Wang and Pengfei Wang and Qiuyue Wang and Yuxuan Wang and Tianbao Xie and Yiheng Xu and Haiyang Xu and Jin Xu and Zhibo Yang and Mingkun Yang and Jianxin Yang and An Yang and Bowen Yu and Fei Zhang and Hang Zhang and Xi Zhang and Bo Zheng and Humen Zhong and Jingren Zhou and Fan Zhou and Jing Zhou and Yuanzhi Zhu and Ke Zhu},
      year={2025},
      eprint={2511.21631},
      archivePrefix={arXiv},
      primaryClass={cs.CV},
      url={https://arxiv.org/abs/2511.21631}, 
}

@inproceedings{wolf2020transformers,

title = {Transformers: State-of-the-Art Natural Language Processing},

author = {Wolf, Thomas and Debut, Lysandre and Sanh, Victor and Chaumond, Julien and Delangue, Clement and Moi, Anthony and Cistac, Pierric and Rault, Tim and Louf, Remi and Funtowicz, Morgan and Davison, Joe and Shleifer, Sam and von Platen, Patrick and Ma, Clara and Jernite, Yacine and Plu, Julien and Xu, Canwen and Le Scao, Teven and Gugger, Sylvain and Drame, Mariama and Lhoest, Quentin and Rush, Alexander M.},

booktitle = {Proceedings of the 2020 Conference on Empirical Methods in Natural Language Processing: System Demonstrations},

pages = {38--45},

year = {2020}

}

@inproceedings{weng2025visnumbench,
  title = {VisNumBench: Evaluating Number Sense of Multimodal Large Language Models},
  author = {Weng, Tengjin and Wang, Jingyi and Jiang, Wenhao and Ming, Zhong},
  booktitle = {Proceedings of the IEEE/CVF International Conference on Computer Vision (ICCV)},
  year = {2025},
  pages = {3830--3840},

}

@misc{alghisi2025vlmcounting,
      title={[De|Re]constructing VLMs' Reasoning in Counting}, 
      author={Simone Alghisi and Gabriel Roccabruna and Massimo Rizzoli and Seyed Mahed Mousavi and Giuseppe Riccardi},
      year={2025},
      eprint={2510.19555},
      archivePrefix={arXiv},
      primaryClass={cs.CV},
      url={https://arxiv.org/abs/2510.19555}, 
}

@misc{hasani2025countingmechanism,
      title={Understanding Counting Mechanisms in Large Language and Vision-Language Models}, 
      author={Hosein Hasani and Amirmohammad Izadi and Fatemeh Askari and Mobin Bagherian and Sadegh Mohammadian and Mohammad Izadi and Mahdieh Soleymani Baghshah},
      year={2025},
      eprint={2511.17699},
      archivePrefix={arXiv},
      primaryClass={cs.CV},
      url={https://arxiv.org/abs/2511.17699}, 
}

@misc{vo2025vlmbiased,
      title={Vision Language Models are Biased}, 
      author={An Vo and Khai-Nguyen Nguyen and Mohammad Reza Taesiri and Vy Tuong Dang and Anh Totti Nguyen and Daeyoung Kim},
      year={2025},
      eprint={2505.23941},
      archivePrefix={arXiv},
      primaryClass={cs.LG},
      url={https://arxiv.org/abs/2505.23941}, 
}

@misc{sengupta2025modest,
      title={Can Vision-Language Models Count? A Synthetic Benchmark and Analysis of Attention-Based Interventions}, 
      author={Saurav Sengupta and Nazanin Moradinasab and Jiebei Liu and Donald E. Brown},
      year={2025},
      eprint={2511.17722},
      archivePrefix={arXiv},
      primaryClass={cs.CV},
      url={https://arxiv.org/abs/2511.17722}, 
}

@misc{koh2024visualwebarena,
      title={VisualWebArena: Evaluating Multimodal Agents on Realistic Visual Web Tasks}, 
      author={Jing Yu Koh and Robert Lo and Lawrence Jang and Vikram Duvvur and Ming Chong Lim and Po-Yu Huang and Graham Neubig and Shuyan Zhou and Ruslan Salakhutdinov and Daniel Fried},
      year={2024},
      eprint={2401.13649},
      archivePrefix={arXiv},
      primaryClass={cs.LG},
      url={https://arxiv.org/abs/2401.13649}, 
}

@misc{lu2024mathvista,
      title={MathVista: Evaluating Mathematical Reasoning of Foundation Models in Visual Contexts}, 
      author={Pan Lu and Hritik Bansal and Tony Xia and Jiacheng Liu and Chunyuan Li and Hannaneh Hajishirzi and Hao Cheng and Kai-Wei Chang and Michel Galley and Jianfeng Gao},
      year={2024},
      eprint={2310.02255},
      archivePrefix={arXiv},
      primaryClass={cs.CV},
      url={https://arxiv.org/abs/2310.02255}, 
}

@misc{liu2023llava,
      title={Visual Instruction Tuning}, 
      author={Haotian Liu and Chunyuan Li and Qingyang Wu and Yong Jae Lee},
      year={2023},
      eprint={2304.08485},
      archivePrefix={arXiv},
      primaryClass={cs.CV},
      url={https://arxiv.org/abs/2304.08485}, 
}

@misc{gekhman2025factualknowledge,
      title={Inside-Out: Hidden Factual Knowledge in LLMs}, 
      author={Zorik Gekhman and Eyal Ben David and Hadas Orgad and Eran Ofek and Yonatan Belinkov and Idan Szpektor and Jonathan Herzig and Roi Reichart},
      year={2025},
      eprint={2503.15299},
      archivePrefix={arXiv},
      primaryClass={cs.CL},
      url={https://arxiv.org/abs/2503.15299}, 
}

@inproceedings{rozanov-rei-2025-stateact,
    title = "{S}tate{A}ct: Enhancing {LLM} Base Agents via Self-prompting and State-tracking",
    author = "Rozanov, Nikolai  and
      Rei, Marek",
    editor = "Kamalloo, Ehsan  and
      Gontier, Nicolas  and
      Lu, Xing Han  and
      Dziri, Nouha  and
      Murty, Shikhar  and
      Lacoste, Alexandre",
    booktitle = "Proceedings of the 1st Workshop for Research on Agent Language Models (REALM 2025)",
    month = jul,
    year = "2025",
    address = "Vienna, Austria",
    publisher = "Association for Computational Linguistics",
    url = "https://aclanthology.org/2025.realm-1.27/",
    doi = "10.18653/v1/2025.realm-1.27",
    pages = "367--385",
    ISBN = "979-8-89176-264-0"
}

@misc{hewitt2019designing,
      title={Designing and Interpreting Probes with Control Tasks}, 
      author={John Hewitt and Percy Liang},
      year={2019},
      eprint={1909.03368},
      archivePrefix={arXiv},
      primaryClass={cs.CL},
      url={https://arxiv.org/abs/1909.03368}, 
}

\appendix

\section{Detailed Results}
\label{Appendix: Detailed Results}

\subsection{SVCCA Details}
\label{Appendix: SVCCA details}
We quantify this using Singular Vector Canonical Correlation Analysis (SVCCA)~\citep{raghu2017svcca}. Consider a pair of trained linear classifiers of the same architecture, represented by weight matrices
\begin{equation}
    W_{\mathrm{gt}} \in \mathbb{R}^{c \times d},
    \qquad
    W_{\mathrm{out}} \in \mathbb{R}^{c \times d},
\end{equation}
where $c$ is the number of classes and $d$ is the hidden dimension. Each row of a probe weight matrix corresponds to a class-specific readout direction, so the row space of the matrix characterizes the subspace used by that probe for prediction.

To extract a stable low-dimensional basis for each probe, we compute the singular value decomposition
\begin{equation}
    W_{\mathrm{gt}}^\top = U_{\mathrm{gt}} \Sigma_{\mathrm{gt}} V_{\mathrm{gt}}^\top,
    \qquad
    W_{\mathrm{out}}^\top = U_{\mathrm{out}} \Sigma_{\mathrm{out}} V_{\mathrm{out}}^\top.
\end{equation}
We then retain the top-$k$ left singular vectors from each decomposition, yielding orthonormal basis matrices
\begin{equation}
    U_{\mathrm{gt}}^{(k)},\, U_{\mathrm{out}}^{(k)} \in \mathbb{R}^{d \times k}.
\end{equation}
These bases span the principal readout subspaces used by the ground-truth and output probes, respectively.

Next, we measure how well these two subspaces align by computing the singular values of their cross-projection matrix:
\begin{equation}
    \rho_1,\dots,\rho_k
    =
    \sigma_1,\dots,\sigma_k
    \left(
        {U_{\mathrm{gt}}^{(k)}}^\top U_{\mathrm{out}}^{(k)}
    \right),
\end{equation}
where each $\rho_i \in [0,1]$ is a canonical correlation. Intuitively, $\rho_i$ measures how similar the $i$-th most aligned directions are across the two probe subspaces. We summarize alignment by the mean canonical correlation:
\begin{equation}
    \mathrm{SVCCA}(W_{\mathrm{gt}}, W_{\mathrm{out}})
    =
    \frac{1}{k}\sum_{i=1}^{k} \rho_i.
\end{equation}

\subsection{Synthetic Datasets Details}
\label{appendix:synthetic_datasets_details}
\begin{table*}[ht]
\centering

\begin{tabular}{ll}
\toprule
\textbf{Property} & \textbf{Values} \\
\midrule
Colors & blue, red, green, yellow, black, orange, purple, cyan, magenta, lime \\
Shapes & circle, square, triangle \\
Color mode & \texttt{sing\_color} (uniform per image), \texttt{diff\_color} (varied per object) \\
Shape mode & \texttt{sing\_shape} (uniform per image), \texttt{diff\_shape} (varied per object) \\
Object count & 1--9 \\
Image resolution & $512 \times 512$ \\
\bottomrule
\end{tabular}
\caption{Properties of the synthetic counting dataset. \texttt{Pillow} python package handles image creation and shape rendering}
\label{tab:dataset-properties}

\end{table*}

\begin{figure}
    \centering
    \includegraphics[width=\linewidth]{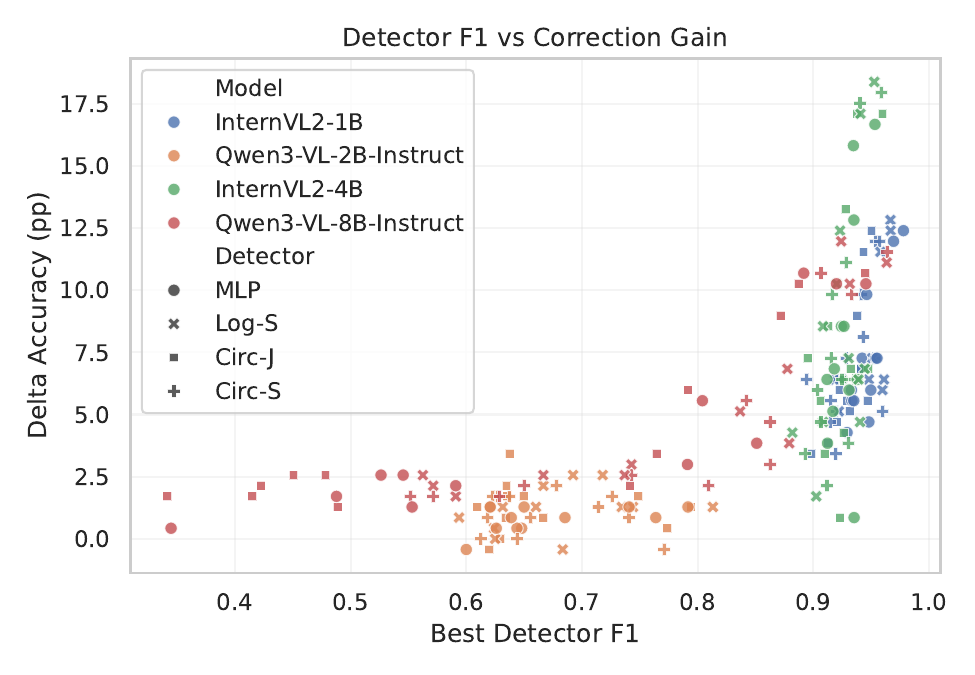}
    \caption{Detector F1 vs.\ correction gain ($\Delta$Acc).
Each point is one run (model $\times$ dataset $\times$
detector $\times$ seed). Spearman $\rho = 0.803$.}
    \label{fig:f1_vs_delta_scatter}
\end{figure}

\begin{figure}[t] 
    \centering
    \begin{subfigure}[b]{1\linewidth} 
        \centering
        \includegraphics[width=\linewidth]{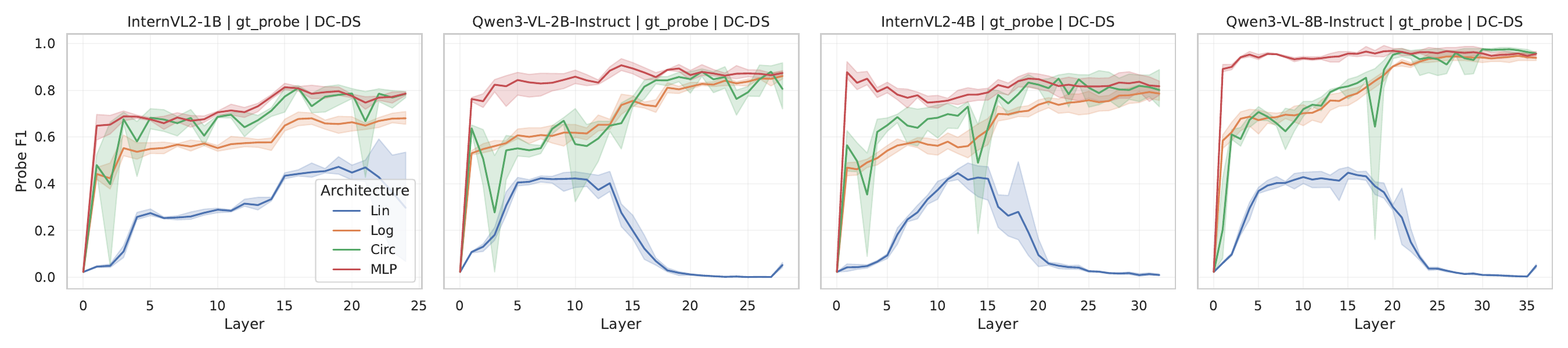}
        \caption{Layer-wise probing F1 score across model depths (ground truth objective).}
    \end{subfigure}
    
    \vspace{0.5cm} 

    \begin{subfigure}[b]{1\linewidth} 
        \centering
        \includegraphics[width=\linewidth]{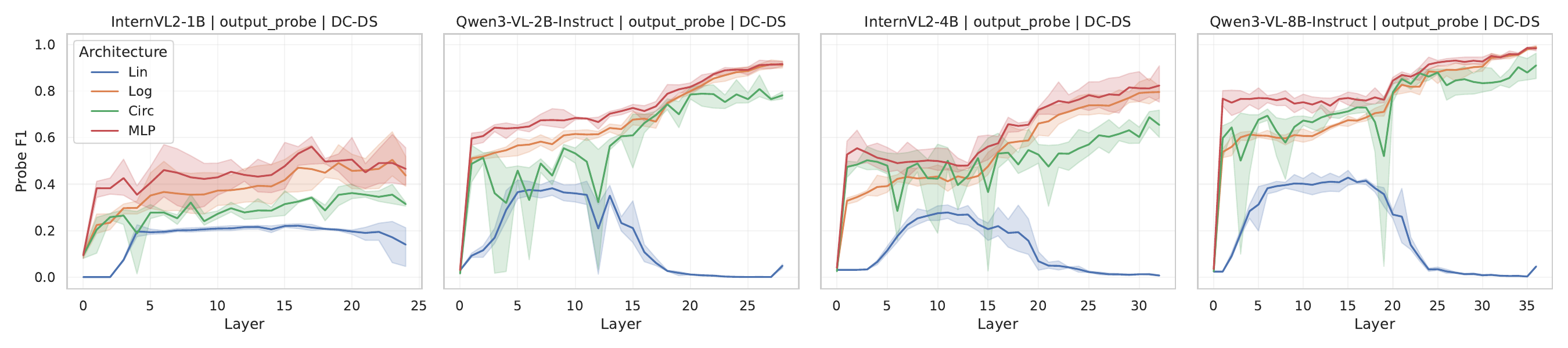}
        \caption{Layer-wise probing F1 score across model depths (model output objective).}
    \end{subfigure}
    
    \vspace{0.5cm} 
    
    \begin{subfigure}[b]{1\linewidth}
        \centering
        \includegraphics[width=\linewidth]{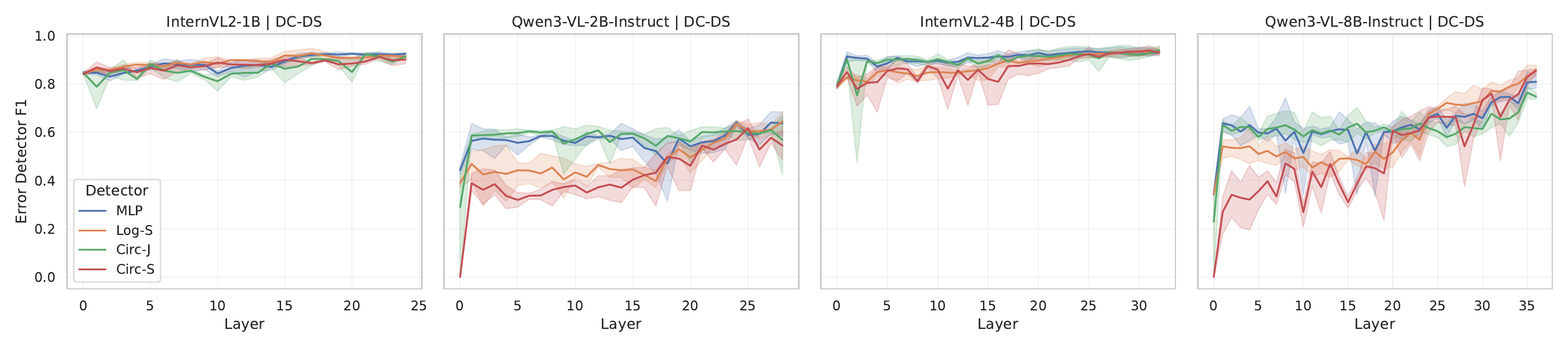}
        \caption{Error detector performance (F1-score) per layer.}
        \label{fig:detector_f1__diff_col_diff_shape}
    \end{subfigure}
    
    \caption{Layer-wise probing and detector performance across model depth. Top: ground-truth probe F1 by layer. Middle: output-supervised probe F1 by layer. Bottom: error-detector F1 by layer. Curves are aggregated across models, and the three random seeds. DC-DS: Different Color Different Shape dataset}
    \label{fig:combined_probing_results__diff_col_diff_shape}
\end{figure}

\begin{figure}[t] 
    \centering
    \begin{subfigure}[b]{1\linewidth} 
        \centering
        \includegraphics[width=\linewidth]{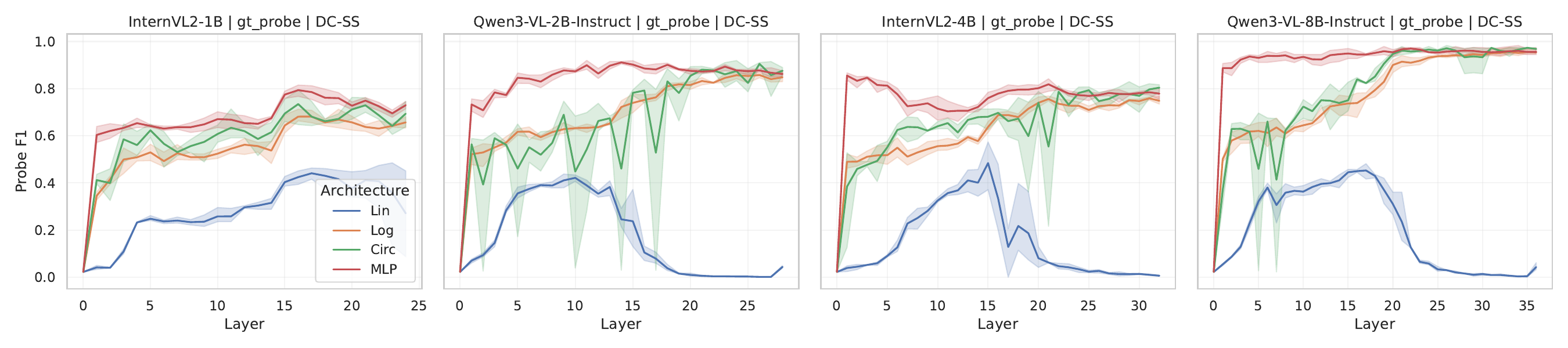}
        \caption{Layer-wise probing F1 score across model depths (ground truth objective).}
    \end{subfigure}
    
    \vspace{0.5cm} 

    \begin{subfigure}[b]{1\linewidth} 
        \centering
        \includegraphics[width=\linewidth]{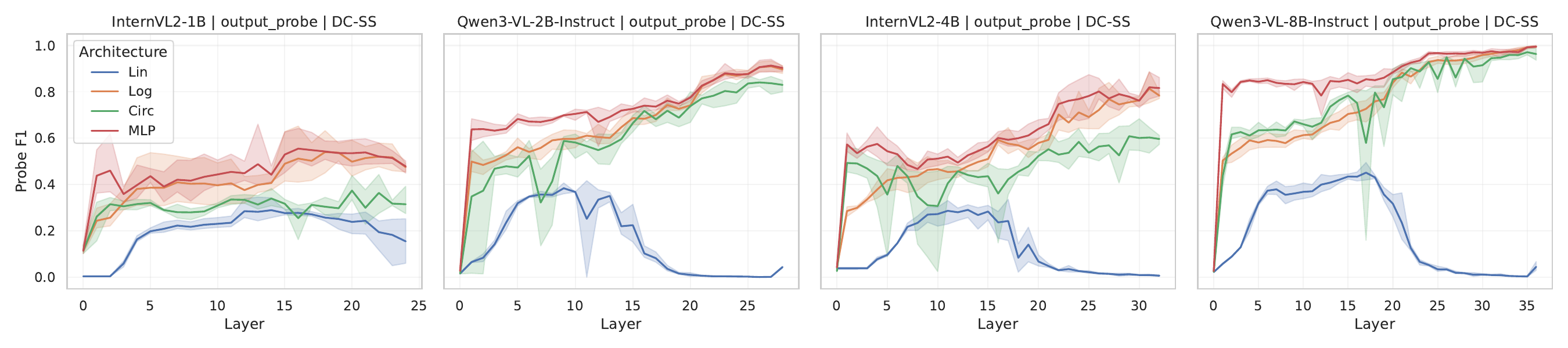}
        \caption{Layer-wise probing F1 score across model depths (model output objective).}
    \end{subfigure}
    
    \vspace{0.5cm} 
    
    \begin{subfigure}[b]{1\linewidth}
        \centering
        \includegraphics[width=\linewidth]{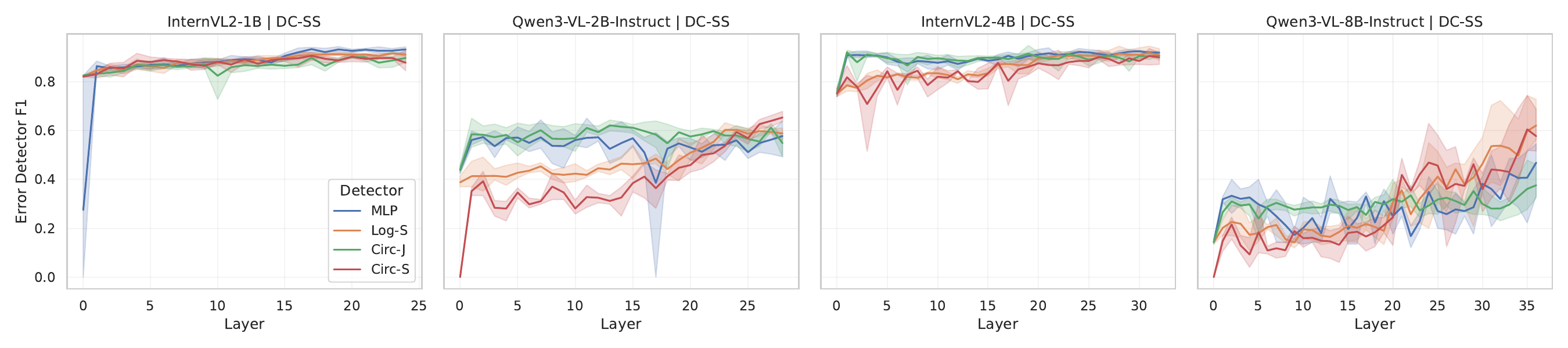}
        \caption{Error detector performance (F1-score) per layer.}
        \label{fig:detector_f1__diff_col_sing_shape}
    \end{subfigure}
    
    \caption{Layer-wise probing and detector performance across model depth. Top: ground-truth probe F1 by layer. Middle: output-supervised probe F1 by layer. Bottom: error-detector F1 by layer. Curves are aggregated across models, and the three random seeds. DC-SS: Different Color Single Shape dataset}
    \label{fig:combined_probing_results__diff_col_sing_shape}
\end{figure}

\begin{figure}[t] 
    \centering
    \begin{subfigure}[b]{1\linewidth} 
        \centering
        \includegraphics[width=\linewidth]{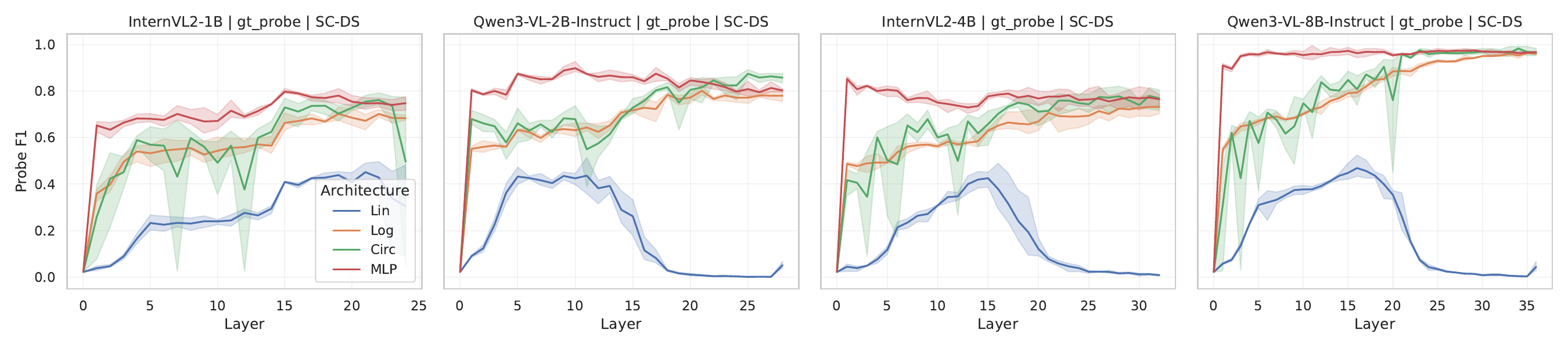}
        \caption{Layer-wise probing F1 score across model depths (ground truth objective).}
    \end{subfigure}
    
    \vspace{0.5cm} 

    \begin{subfigure}[b]{1\linewidth} 
        \centering
        \includegraphics[width=\linewidth]{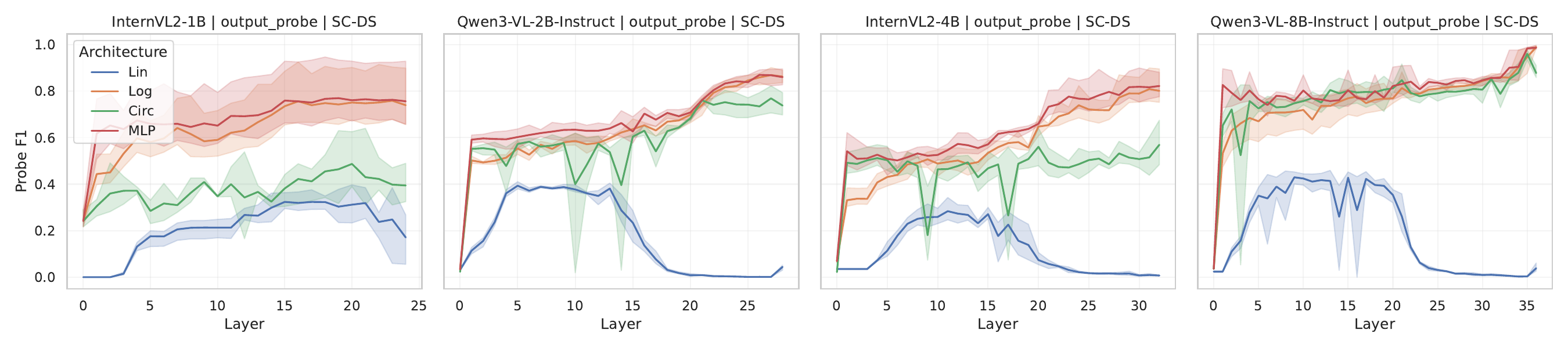}
        \caption{Layer-wise probing F1 score across model depths (model output objective).}
    \end{subfigure}
    
    \vspace{0.5cm} 
    
    \begin{subfigure}[b]{1\linewidth}
        \centering
        \includegraphics[width=\linewidth]{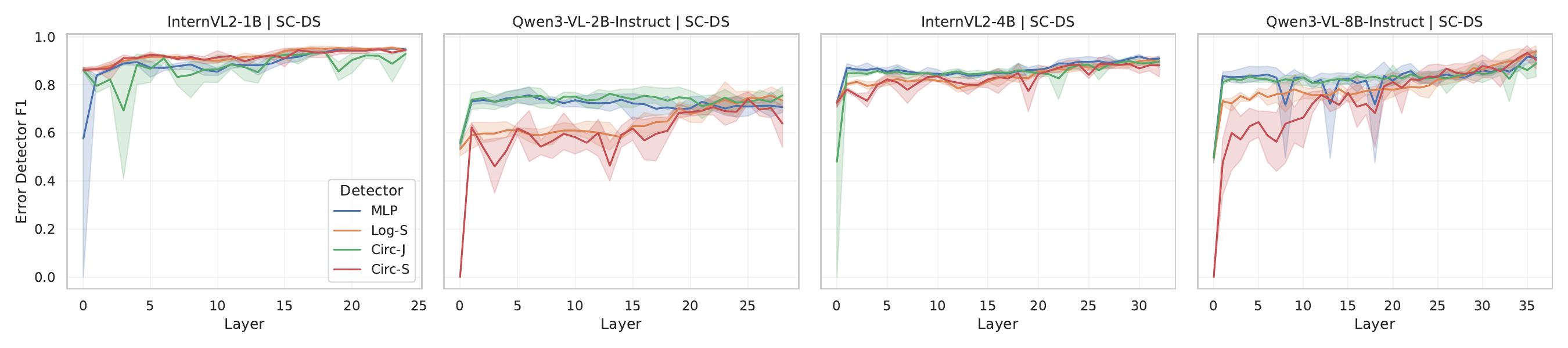}
        \caption{Error detector performance (F1-score) per layer.}
        \label{fig:detector_f1__sing_col_diff_shape}
    \end{subfigure}
    
    \caption{Layer-wise probing and detector performance across model depth. Top: ground-truth probe F1 by layer. Middle: output-supervised probe F1 by layer. Bottom: error-detector F1 by layer. Curves are aggregated across models, and the three random seeds. SC-DS: Single Color Different Shape dataset.}
    \label{fig:combined_probing_results__sing_col_diff_shape}
\end{figure}

\begin{figure}[t] 
    \centering
    \begin{subfigure}[b]{1\linewidth} 
        \centering
        \includegraphics[width=\linewidth]{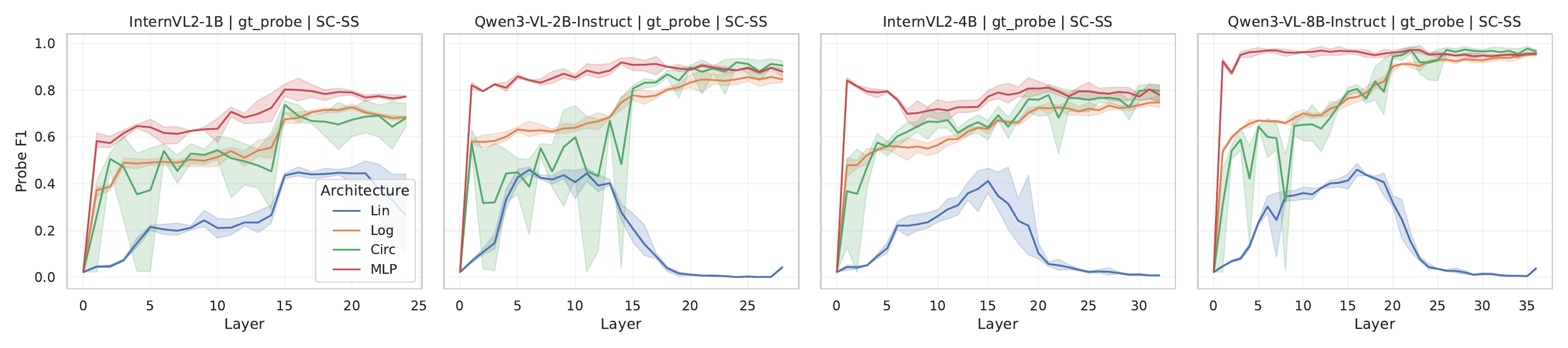}
        \caption{Layer-wise probing F1 score across model depths (ground truth objective).}
    \end{subfigure}
    
    \vspace{0.5cm} 

    \begin{subfigure}[b]{1\linewidth} 
        \centering
        \includegraphics[width=\linewidth]{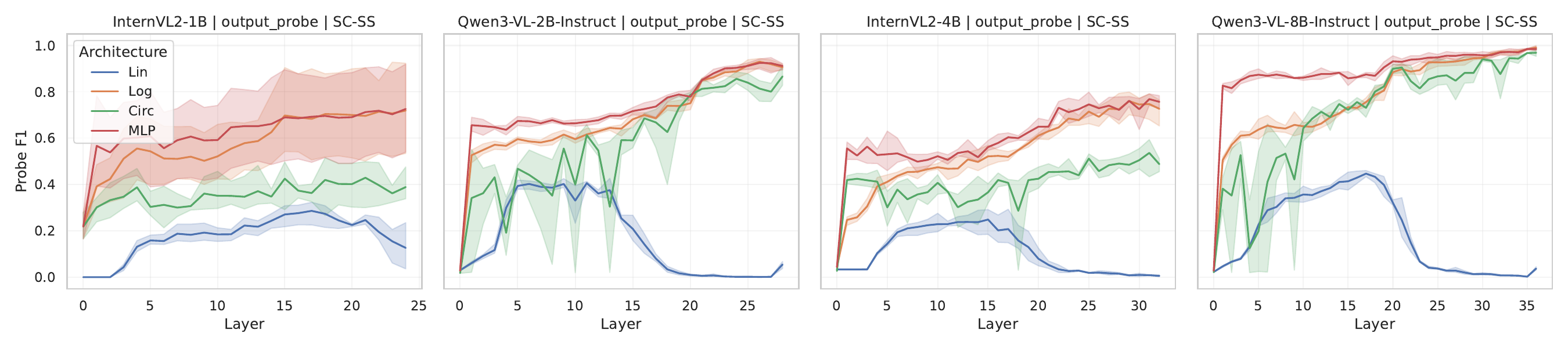}
        \caption{Layer-wise probing F1 score across model depths (model output objective).}
    \end{subfigure}
    
    \vspace{0.5cm} 
    
    \begin{subfigure}[b]{1\linewidth}
        \centering
        \includegraphics[width=\linewidth]{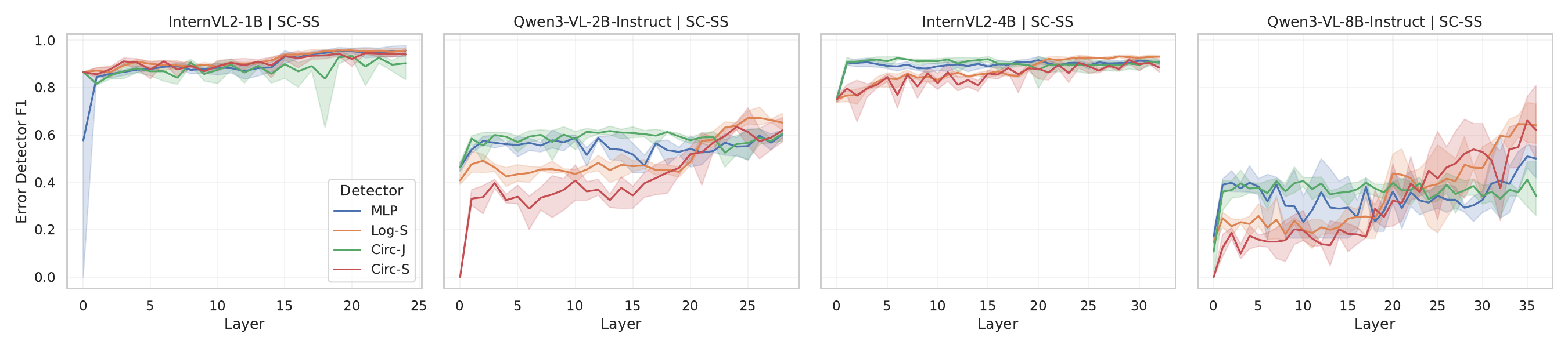}
        \caption{Error detector performance (F1-score) per layer.}
        \label{fig:detector_f1__sing_col_sing_shape}
    \end{subfigure}
    
    \caption{Layer-wise probing and detector performance across model depth. Top: ground-truth probe F1 by layer. Middle: output-supervised probe F1 by layer. Bottom: error-detector F1 by layer. Curves are aggregated across models, and the three random seeds. SC-SS: Single Color Single Shape dataset.}
    \label{fig:combined_probing_results__sing_col_sing_shape}
\end{figure}

\begin{figure}[t] 
    \centering
    \begin{subfigure}[b]{1\linewidth} 
        \centering
        \includegraphics[width=\linewidth]{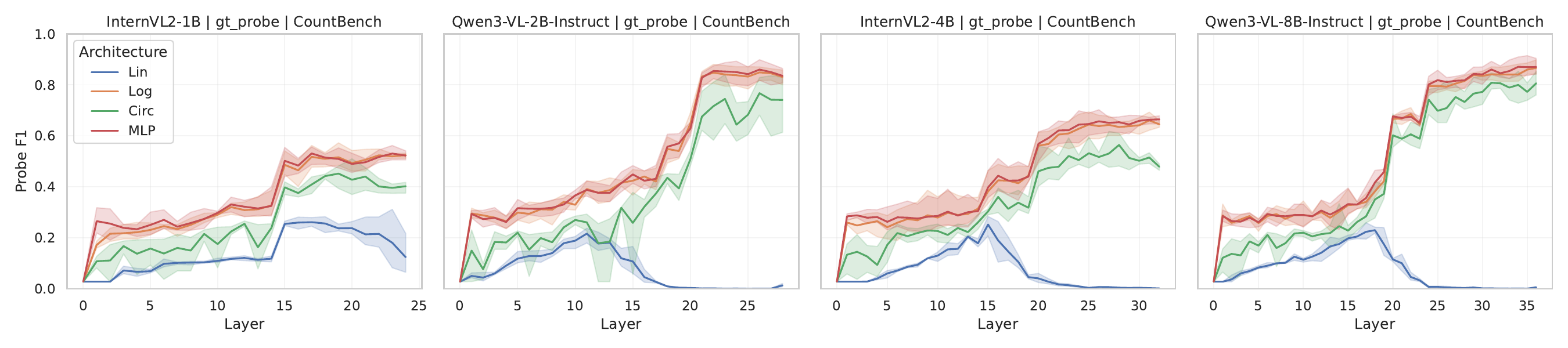}
        \caption{Layer-wise probing F1 score across model depths (ground truth objective).}
    \end{subfigure}
    
    \vspace{0.5cm} 

    \begin{subfigure}[b]{1\linewidth} 
        \centering
        \includegraphics[width=\linewidth]{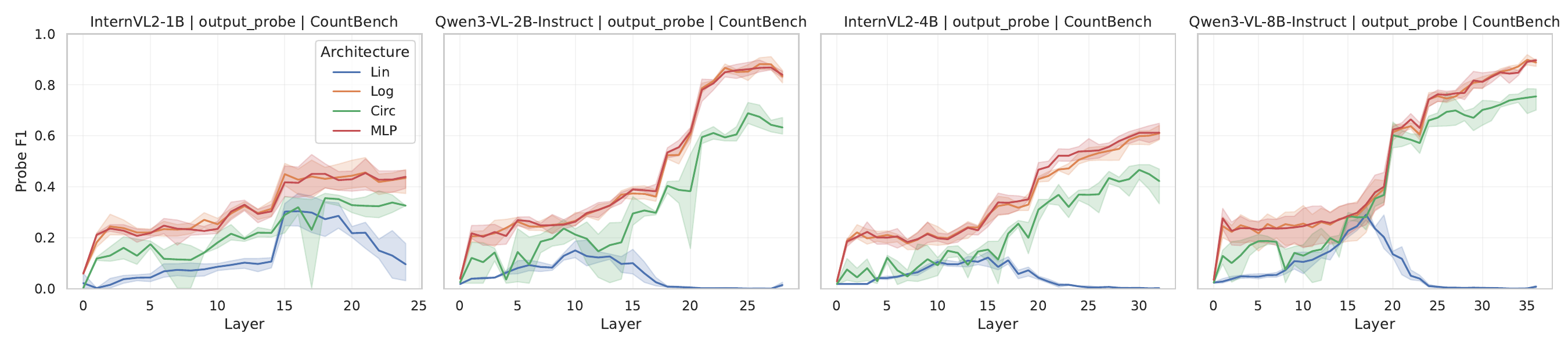}
        \caption{Layer-wise probing F1 score across model depths (model output objective).}
    \end{subfigure}
    
    \vspace{0.5cm} 
    
    \begin{subfigure}[b]{1\linewidth}
        \centering
        \includegraphics[width=\linewidth]{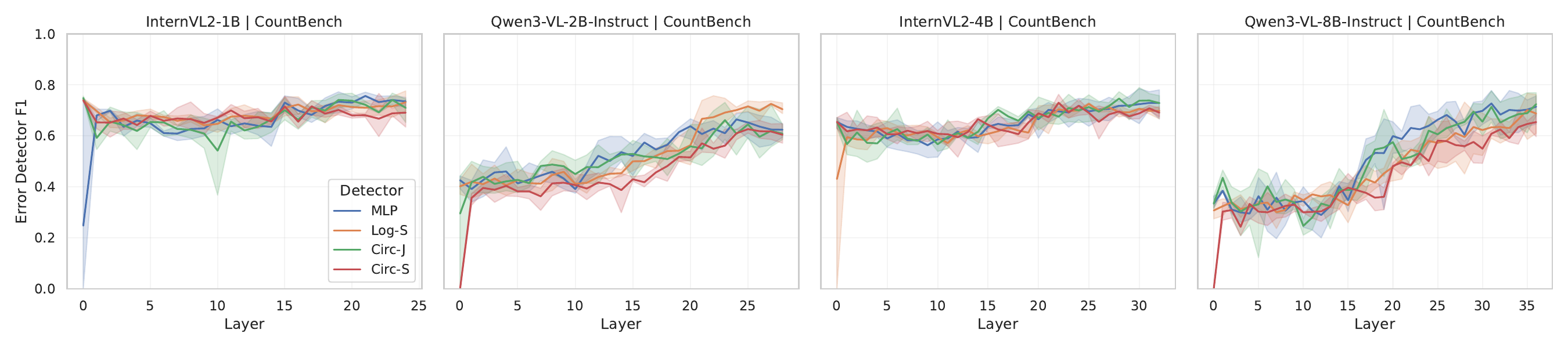}
        \caption{Error detector performance (F1-score) per layer.}
        \label{fig:detector_f1__countbench}
    \end{subfigure}
    
    \caption{Layer-wise probing and detector performance across model depth. Top: ground-truth probe F1 by layer. Middle: output-supervised probe F1 by layer. Bottom: error-detector F1 by layer. Curves are aggregated across models, and the three random seeds. CountBench dataset}
    \label{fig:combined_probing_results__countbench}
\end{figure}

\begin{table}[t]
\centering
\scriptsize
\setlength{\tabcolsep}{4pt}
\begin{tabular}{lccccc}
\toprule
\textbf{Dataset} & \textbf{Model} & \textbf{MLP} & \textbf{Log-S} & \textbf{Circ-J} & \textbf{Circ-S} \\
\midrule
\multirow{4}{*}{\texttt{DC-DS}}
& InternVL2-1B & +5.27 & \textbf{\textcolor{posgreen}{+6.41}} & +5.13 & +4.42 \\
& InternVL2-4B & +15.10 & \textbf{\textcolor{posgreen}{+15.95}} & +15.81 & +15.53 \\
& Qwen3-VL-2B  & +0.57 & +0.43 & \textbf{\textcolor{posgreen}{+0.57}} & \textbf{\textcolor{posgreen}{+0.57}} \\
& Qwen3-VL-8B  & +4.13 & \textbf{\textcolor{posgreen}{+5.27}} & +3.85 & +4.42 \\

\midrule
\multirow{4}{*}{\texttt{DC-SS}}
& InternVL2-1B & +5.84 & +5.98 & +4.42 & \textbf{\textcolor{posgreen}{+6.41}} \\
& InternVL2-4B & +4.13 & \textbf{\textcolor{posgreen}{+4.42}} & +3.70 & +3.99 \\
& Qwen3-VL-2B  & +0.57 & +0.71 & +1.00 & \textbf{\textcolor{posgreen}{+1.28}} \\
& Qwen3-VL-8B  & +1.85 & \textbf{\textcolor{posgreen}{+2.56}} & +1.99 & +1.99 \\

\midrule
\multirow{4}{*}{\texttt{SC-DS}}
& InternVL2-1B & \textbf{\textcolor{posgreen}{+6.84}} & +6.55 & +6.41 & \textbf{\textcolor{posgreen}{+6.84}} \\
& InternVL2-4B & \textbf{\textcolor{posgreen}{+7.83}} & +7.69 & +7.12 & +7.69 \\
& Qwen3-VL-2B  & +1.14 & \textbf{\textcolor{posgreen}{+1.28}} & +1.14 & +0.71 \\
& Qwen3-VL-8B  & +10.40 & \textbf{\textcolor{posgreen}{+11.11}} & +9.97 & +10.68 \\

\midrule
\multirow{4}{*}{\texttt{SC-SS}}
& InternVL2-1B & +11.40 & \textbf{\textcolor{posgreen}{+12.25}} & +10.97 & +11.25 \\
& InternVL2-4B & +5.56 & \textbf{\textcolor{posgreen}{+6.13}} & +5.84 & +4.99 \\
& Qwen3-VL-2B  & +1.00 & \textbf{\textcolor{posgreen}{+2.42}} & +2.28 & +1.57 \\
& Qwen3-VL-8B  & +1.71 & \textbf{\textcolor{posgreen}{+2.28}} & +1.99 & +1.99 \\
\bottomrule
\end{tabular}
\caption{Stratified detector-guided gains: mean $\Delta$Acc (pp) for \textbf{Ours} over \textbf{Before}, computed per model/dataset and averaged over 3 seeds. Detector abbreviations are \texttt{Log-S} (Logistic-Separately), \texttt{Circ-J} (Circular-Jointly), and \texttt{Circ-S} (Circular-Separately). Bold and green mark the best detector(s) per row.}
\label{tab:detectors-stratified}
\end{table}

\begin{figure}
    \centering
    \includegraphics[width=\linewidth]{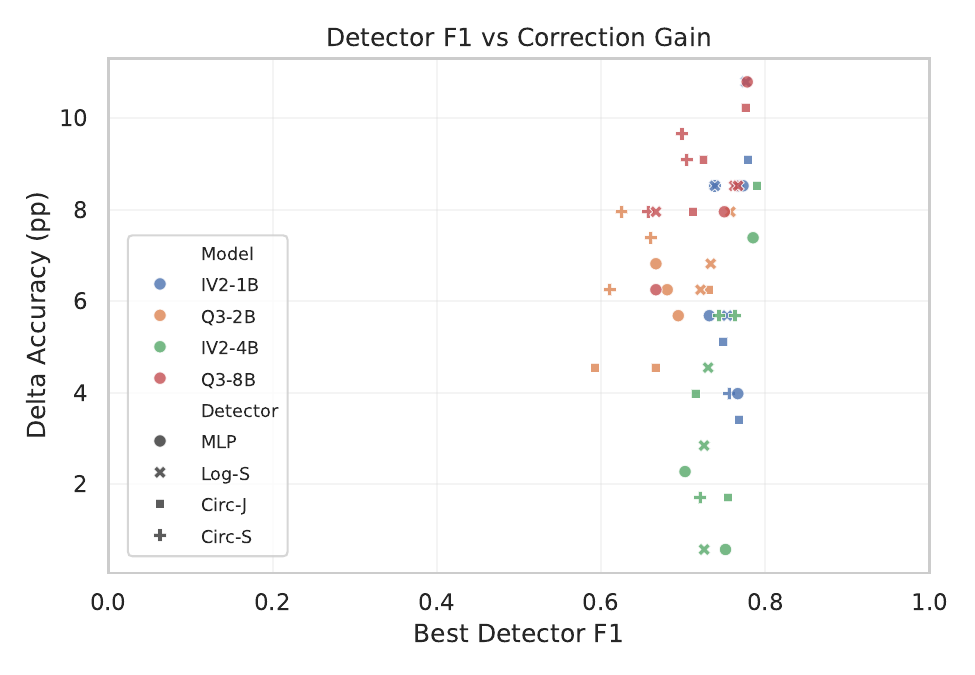}
    \caption{Detector F1 vs.\ correction gain ($\Delta$Acc).
Each point is one run. The dataset is CountBench }
    \label{fig:f1_vs_delta_scatter_countbench}
\end{figure}

\begin{table*}[t!]
\centering
\scriptsize
\setlength{\tabcolsep}{3pt} 
\begin{tabular}{ll ccc ccc ccc ccc}
\toprule
& & \multicolumn{3}{c}{\textbf{MLP}} & \multicolumn{3}{c}{\textbf{Log-S}} & \multicolumn{3}{c}{\textbf{Circ-J}} & \multicolumn{3}{c}{\textbf{Circ-S}} \\
\cmidrule(lr){3-5} \cmidrule(lr){6-8} \cmidrule(lr){9-11} \cmidrule(lr){12-14}
\textbf{Data} & \textbf{Model} & \tiny{TP} & \tiny{FP} & \tiny{$\Delta$Acc (pp)} & \tiny{TP} & \tiny{FP} & \tiny{$\Delta$Acc (pp)} & \tiny{TP} & \tiny{FP} & \tiny{$\Delta$Acc (pp)} & \tiny{TP} & \tiny{FP} & \tiny{$\Delta$Acc (pp)} \\
\midrule

\multirow{4}{*}{\texttt{DC-DS}} 
& IVL2-1B & 8.9 & 86.6 & 5.26 & 10.1 & 88.6 & 6.41 & 8.8 & 85.1 & 5.13 & 8.4 & 79.1 & 4.40 \\
& IVL2-4B & 25.9 & 78.7 & 15.09 & 27.5 & 76.7 & 15.94 & 26.5 & 83.8 & 15.81 & 26.2 & 80.0 & 15.51 \\

& Q3-2B   & 4.2 & 97.6 & 0.56 & 5.3 & 95.0 & 0.43 & 4.4 & 98.1 & 0.56 & 5.5 & 95.0 & 0.56 \\
& Q3-8B   & 25.1 & 100.0 & 4.15 & 30.5 & 88.9 & 5.26 & 26.0 & 93.0 & 3.85 & 26.0 & 93.3 & 4.40 \\

\midrule

\multirow{4}{*}{\texttt{DC-SS}} 
& IVL2-1B & 9.3 & 91.7 & 5.85 & 9.4 & 96.3 & 5.98 & 7.9 & 89.4 & 4.40 & 10.0 & 97.8 & 6.41 \\
& IVL2-4B & 9.9 & 65.3 & 4.15 & 10.6 & 66.4 & 4.40 & 9.1 & 75.5 & 3.72 & 9.3 & 74.5 & 3.97 \\

& Q3-2B   & 4.2 & 97.3 & 0.56 & 6.1 & 97.4 & 0.73 & 5.3 & 98.4 & 0.98 & 7.7 & 100.0 & 1.28 \\
& Q3-8B   & 48.9 & 100.0 & 1.84 & 61.0 & 100.0 & 2.56 & 42.2 & 99.3 & 2.01 & 48.3 & 83.3 & 2.01 \\

\midrule

\multirow{4}{*}{\texttt{SC-DS}} 
& IVL2-1B & 11.1 & 62.3 & 6.84 & 10.5 & 61.1 & 6.54 & 11.0 & 71.2 & 6.41 & 10.7 & 73.5 & 6.84 \\
& IVL2-4B & 17.1 & 66.8 & 7.82 & 18.1 & 61.5 & 7.69 & 16.3 & 75.9 & 7.14 & 17.1 & 77.8 & 7.69 \\

& Q3-2B   & 5.0 & 96.4 & 1.15 & 7.3 & 91.0 & 1.28 & 5.2 & 95.9 & 1.15 & 4.6 & 92.1 & 0.73 \\
& Q3-8B   & 34.2 & 96.7 & 10.38 & 35.5 & 100.0 & 11.11 & 34.0 & 93.0 & 9.96 & 34.9 & 93.3 & 10.68 \\

\midrule

\multirow{4}{*}{\texttt{SC-SS}} 
& IVL2-1B & 16.0 & 86.1 & 11.41 & 16.5 & 100.0 & 12.26 & 16.0 & 85.0 & 10.98 & 16.0 & 92.1 & 11.24 \\
& IVL2-4B & 11.9 & 77.3 & 5.56 & 12.8 & 72.6 & 6.11 & 12.2 & 75.7 & 5.85 & 10.9 & 76.4 & 5.00 \\

& Q3-2B   & 9.5 & 93.4 & 0.98 & 14.7 & 94.0 & 2.44 & 13.2 & 95.4 & 2.26 & 12.4 & 92.0 & 1.58 \\
& Q3-8B   & 33.8 & 100.0 & 1.71 & 42.7 & 100.0 & 2.26 & 34.9 & 99.0 & 2.01 & 41.3 & 94.4 & 2.01 \\

\bottomrule
\end{tabular}
\caption{Full detector analysis by model and dataset, averaged over the three seeds. Within each detector block, columns report \textbf{TP} (true-positive correction rate, \%), \textbf{FP} (false-positive preservation rate, \%), and \textbf{$\Delta$Acc} (net accuracy gain in percentage points). Model abbreviations are \textbf{IVL2} for InternVL2 and \textbf{Q3} for Qwen3-VL; detector abbreviations are \texttt{Log-S} (Logistic-Separately), \texttt{Circ-J} (Circular-Jointly), and \texttt{Circ-S} (Circular-Separately).}
\label{tab:full_detector_analysis}
\end{table*}

\begin{figure*}
    \centering
    \begin{subfigure}[b]{0.48\textwidth}
        \centering
        \includegraphics[width=\linewidth]{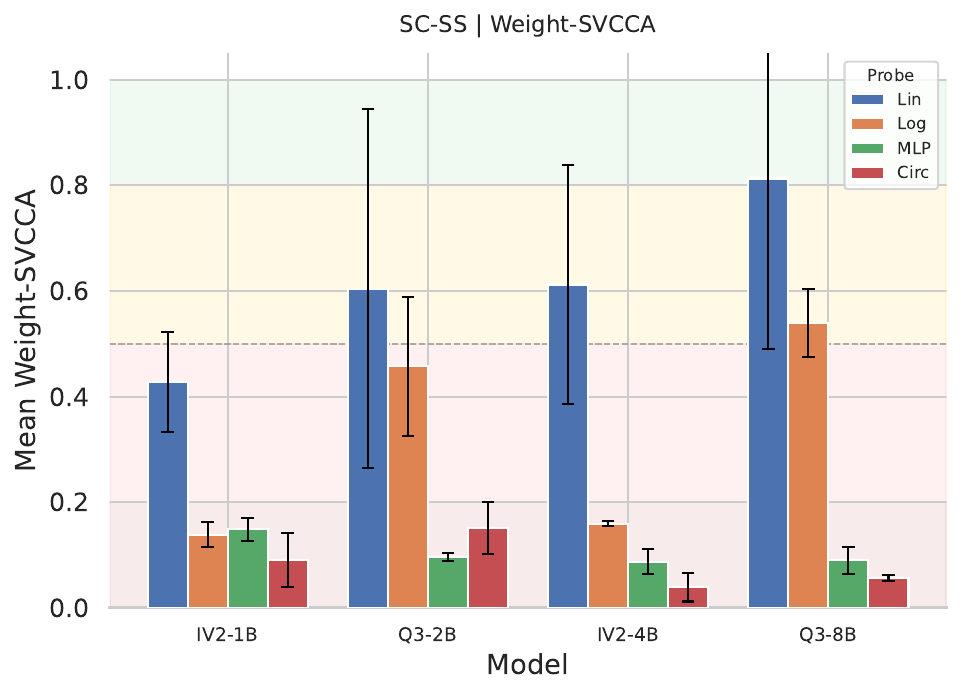}
        \caption{Single color single shape}
        \label{fig:svcca Single color single shape}
    \end{subfigure}
    \hfill
    \begin{subfigure}[b]{0.48\textwidth}
        \centering
        \includegraphics[width=\linewidth]{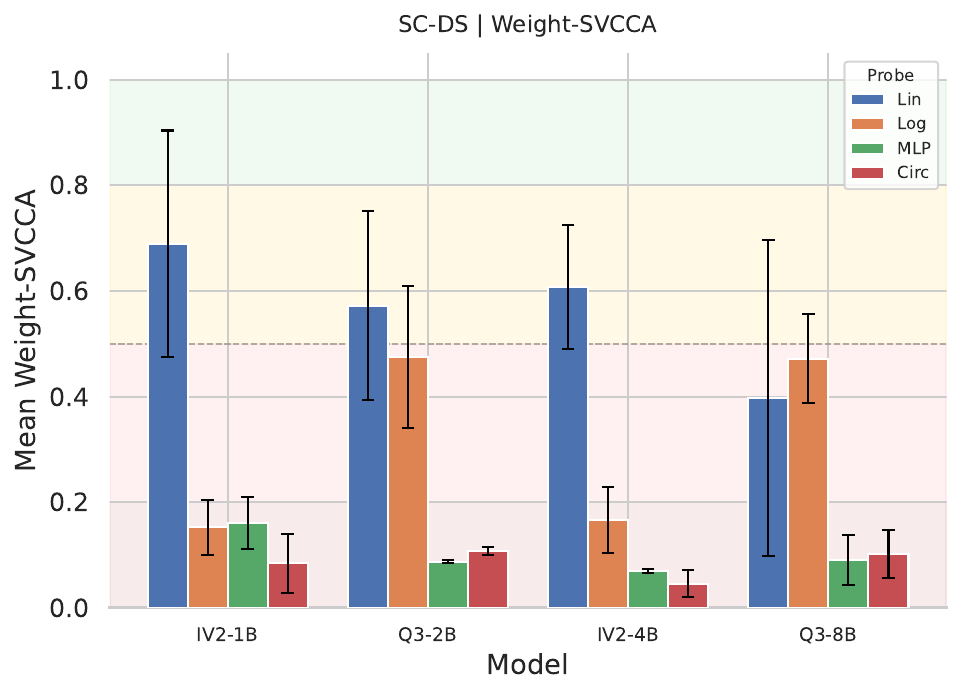}
        \caption{Single color different shape}
        \label{fig:svcca Single color different shape}
    \end{subfigure}

    \vspace{0.3cm} 

    \begin{subfigure}[b]{0.48\textwidth}
        \centering
        \includegraphics[width=\linewidth]{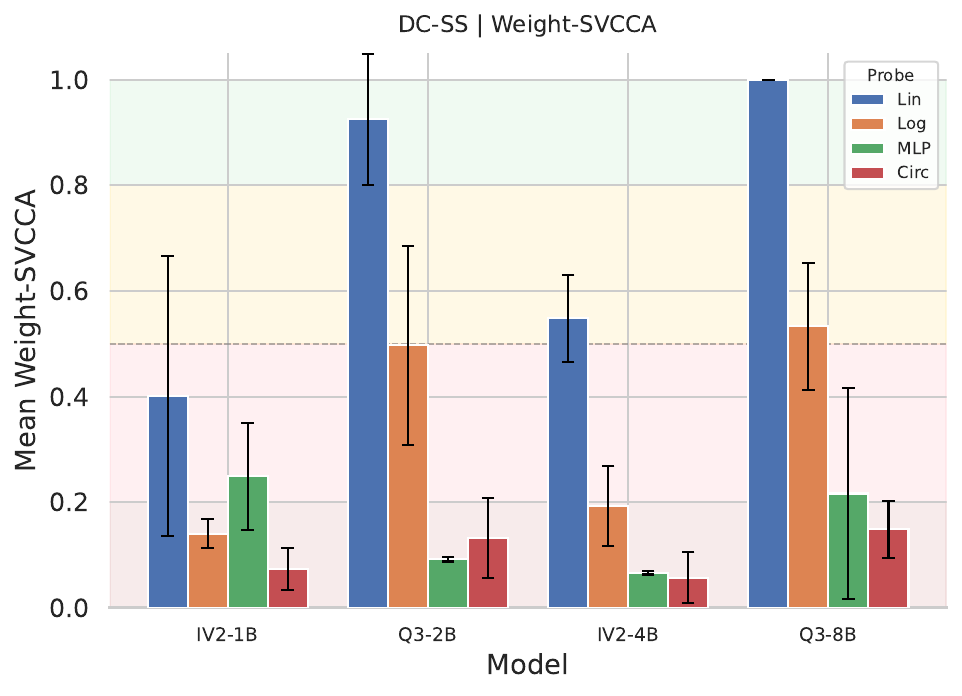}
        \caption{Different color single shape}
        \label{fig:svcca Different color single shape}
    \end{subfigure}
    \hfill
    \begin{subfigure}[b]{0.48\textwidth}
        \centering
        \includegraphics[width=\linewidth]{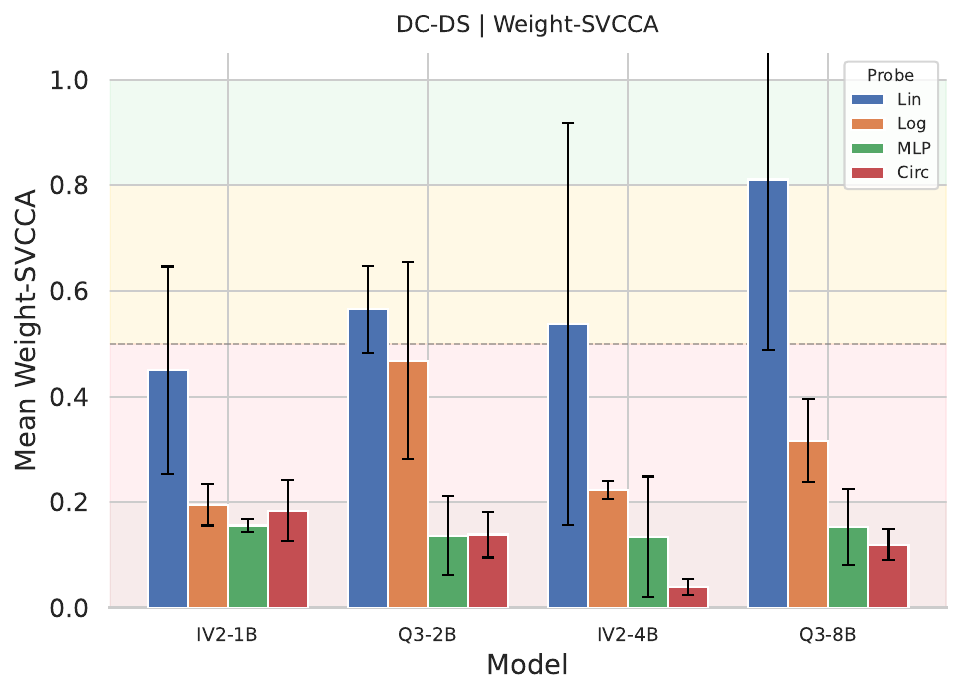}
        \caption{Different color different shape}
        \label{fig:svcca Different color different shape}
    \end{subfigure}
    
    \caption{Layer-wise SVCCA similarity scores for various Vision Language Models and Datasets. High scores indicate strong alignment between ground truth and output probes.}
    \label{fig:svcca_per_dataset}

\end{figure*}

\begin{figure*}[t] 
    \centering
    
    \begin{subfigure}[b]{0.48\linewidth} 
        \centering
        \includegraphics[width=\linewidth]{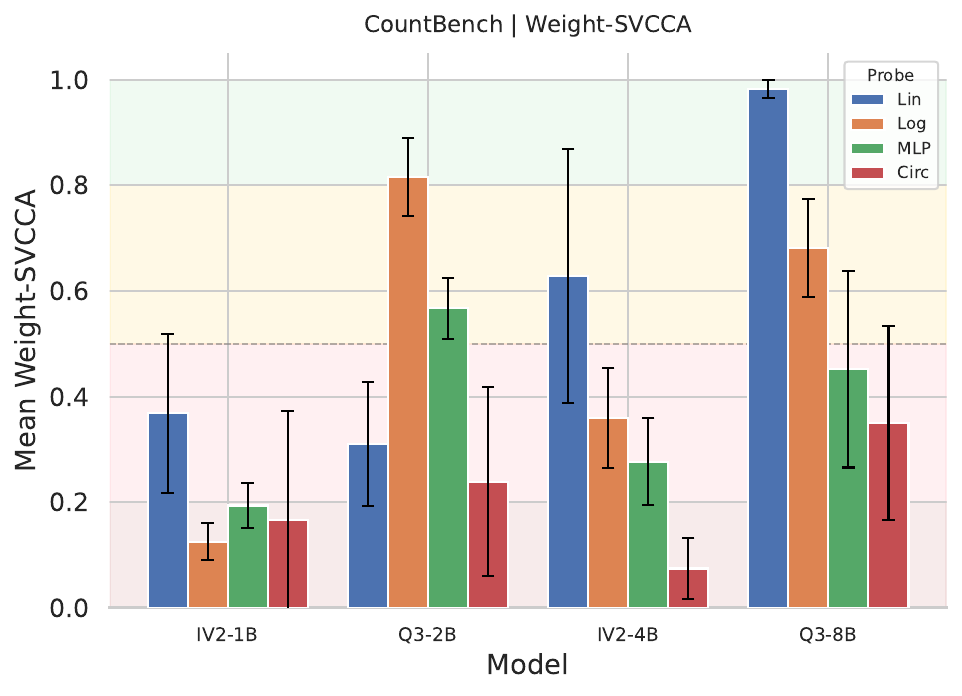}
        \caption{Mean weight-SVCCA by probe family.}
        \label{fig:svcca_by_probe_type_weight_countbench}
    \end{subfigure}
    \hfill 
    \begin{subfigure}[b]{0.48\linewidth} 
        \centering
        \includegraphics[width=\linewidth]{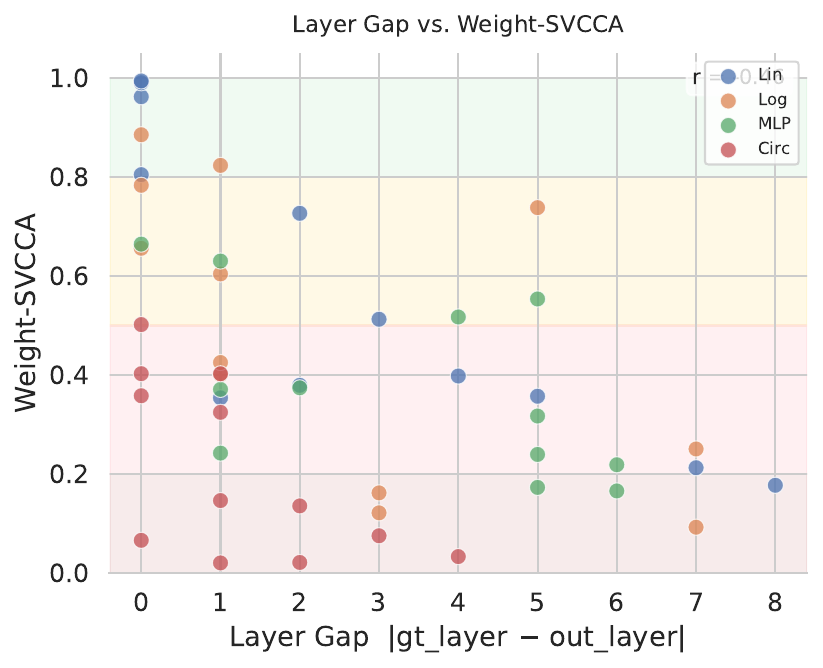}
        \caption{Weight-SVCCA versus best-layer gap.}
        \label{fig:svcca_layer_gap_scatter_weight_countbench}
    \end{subfigure}
    
    \caption{Weight-space SVCCA analysis of \texttt{gt\_probe} versus \texttt{output\_probe}. (\textbf{Left}) Mean SVCCA by probe family across the full 4-model, 3-seed sweep. MLP and Circular probes lie in the near-orthogonal regime, while Linear probes show substantially higher alignment. (\textbf{Right}) SVCCA versus best-layer gap $|l_{gt}-l_{out}|$. Larger layer disagreement tends to coincide with lower alignment, but low SVCCA persists even at small layer gaps, indicating that subspace divergence is not reducible to depth mismatch alone. CountBench Dataset}
    \label{fig:combined_wsvcca_results_countbench}
\end{figure*}

\begin{figure*}[t] 
    \centering
    
    \begin{subfigure}[b]{0.45\linewidth} 
        \centering
        \includegraphics[width=\linewidth]{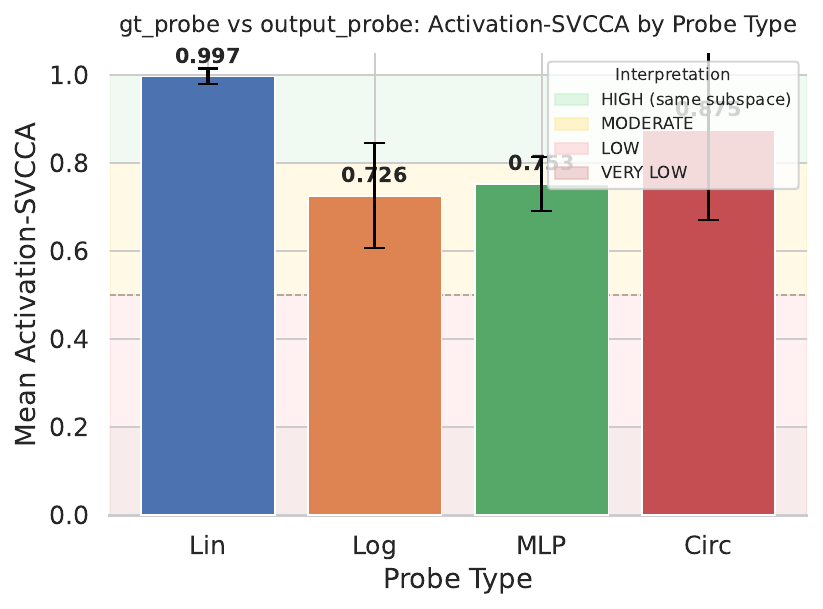}
        \caption{Mean activation-SVCCA by probe family.}
        \label{fig:svcca_by_probe_type_activation}
    \end{subfigure}
    \hfill 
    \begin{subfigure}[b]{0.45\linewidth} 
        \centering
        \includegraphics[width=\linewidth]{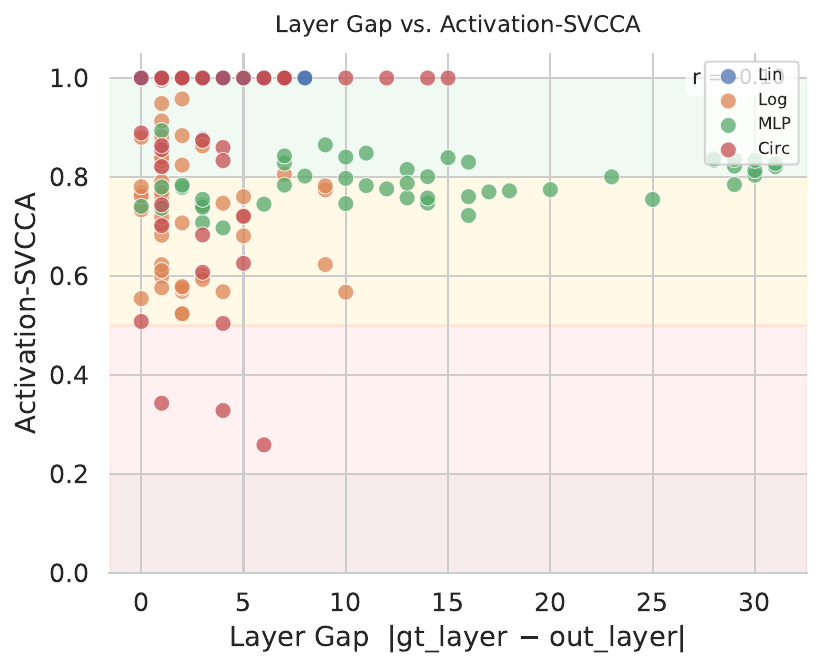}
        \caption{Activation-SVCCA versus best-layer gap.}
        \label{fig:svcca_layer_gap_scatter_activation}
    \end{subfigure}
    
    \caption{Activation-space SVCCA analysis of \texttt{gt\_probe} versus \texttt{output\_probe}. }
    \label{fig:combined_asvcca_results}
\end{figure*}

\section{Correction Prompt Template}
\label{app:correction_prompt}

The following prompt is used during the second pass of the detector-guided
self-correction pipeline (\S\ref{sec:self correction}). 

\begin{tcolorbox}[
  colback=gray!5,
  colframe=gray!50,
  boxrule=0.5pt,
  arc=3pt,
  left=8pt, right=8pt, top=6pt, bottom=6pt,
  title={\texttt{$f_{\mathrm{corr}}(q,\, \hat{y}^{(1)})$}},
  fonttitle=\small\bfseries
]
\small
\texttt{<image>}\\[4pt]
\texttt{<q>: How many object(s) are in this image? Answer:}\\[4pt]
\texttt{Your previous answer was: }$\hat{y}^{(1)}$\\[4pt]
\texttt{That's definitely wrong. The correct count should be:}
\end{tcolorbox}

\noindent The system prompt used for all passes is:

\begin{tcolorbox}[
  colback=gray!5,
  colframe=gray!50,
  boxrule=0.5pt,
  arc=3pt,
  left=8pt, right=8pt, top=6pt, bottom=6pt,
  title={\texttt{System Prompt}},
  fonttitle=\small\bfseries
]
\small
\texttt{You are a helpful assistant. Output only the final
count in the format <count>.}
\end{tcolorbox}


\subsection{Experiments on Additional Prompts}
\label{appendix:prompt_comparison_internvl1b}

To evaluate the sensitivity of detector-guided self-correction to the specific framing of the intervention prompt, we compare three distinct reprompting strategies. The exact text for each prompt variation is defined as follows:

\begin{tcolorbox}[
  colback=gray!5,
  colframe=gray!50,
  boxrule=0.5pt,
  arc=3pt,
  left=8pt, right=8pt, top=6pt, bottom=6pt,
  title={\texttt{stronger}},
  fonttitle=\small\bfseries
]
\small
\texttt{That's definitely wrong. The correct count should be:}
\end{tcolorbox}

\begin{tcolorbox}[
  colback=gray!5,
  colframe=gray!50,
  boxrule=0.5pt,
  arc=3pt,
  left=8pt, right=8pt, top=6pt, bottom=6pt,
  title={\texttt{strongv2}},
  fonttitle=\small\bfseries
]
\small
\texttt{That's incorrect. Recount every visible item in the image carefully.}
\end{tcolorbox}

\begin{tcolorbox}[
  colback=gray!5,
  colframe=gray!50,
  boxrule=0.5pt,
  arc=3pt,
  left=8pt, right=8pt, top=6pt, bottom=6pt,
  title={\texttt{counting}},
  fonttitle=\small\bfseries
]
\small
\texttt{I made an error in my previous count—there appears to be a different number of objects in the image. I'll recount carefully, step by step:}
\end{tcolorbox}

As shown in Table~\ref{tab:internvl1b_prompt_comparison}, the \texttt{counting} prompt yields the highest absolute gains on three of the four datasets, peaking at a $+16.67$ improvement on \texttt{sing\_col\_diff\_shape}. However, \texttt{stronger} remains the most conservative and stable approach. When evaluated by preservation-oriented metrics, \texttt{stronger} maintains the highest average false-positive preservation rates (ranging from $71.95\%$ to $95.21\%$ across the synthetic datasets) and achieves the highest combined effectiveness on a majority of splits. This highlights a clear architectural trade-off between aggressive error correction and the preservation of baseline coherence.

\begin{table*}[t]
\centering
\resizebox{\textwidth}{!}{
\begin{tabular}{l c ccc ccc ccc}
\toprule
& & \multicolumn{3}{c}{\textbf{stronger}} & \multicolumn{3}{c}{\textbf{strongv2}} & \multicolumn{3}{c}{\textbf{counting}} \\
\cmidrule(lr){3-5} \cmidrule(lr){6-8} \cmidrule(lr){9-11}
\textbf{Dataset} & \textbf{Raw} & \textbf{Naive} & \textbf{Random-K} & \textbf{Ours ($\Delta$)} & \textbf{Naive} & \textbf{Random-K} & \textbf{Ours ($\Delta$)} & \textbf{Naive} & \textbf{Random-K} & \textbf{Ours ($\Delta$)} \\
\midrule
\texttt{diff\_col\_diff\_shape} & 27.35 & \textcolor{negred}{22.65} & \textcolor{negred}{22.54} & 32.05 (\textcolor{posgreen}{+4.70}) & \textcolor{negred}{26.92} & \textcolor{negred}{25.43} & 30.77 (\textcolor{posgreen}{+3.42}) & \textcolor{negred}{22.65} & \textcolor{negred}{22.97} & \textbf{36.01} (\textcolor{posgreen}{+8.65}) \\
\texttt{diff\_col\_sing\_shape} & 29.49 & \textcolor{negred}{27.78} & \textcolor{negred}{26.71} & 36.22 (\textcolor{posgreen}{+6.73}) & 30.34 & \textcolor{negred}{27.46} & 36.64 (\textcolor{posgreen}{+7.16}) & \textcolor{negred}{23.93} & \textcolor{negred}{23.72} & \textbf{37.71} (\textcolor{posgreen}{+8.23}) \\
\texttt{sing\_col\_diff\_shape} & 22.65 & \textcolor{negred}{22.22} & \textcolor{negred}{20.94} & 29.59 (\textcolor{posgreen}{+6.94}) & \textcolor{negred}{19.23} & \textcolor{negred}{17.52} & 26.71 (\textcolor{posgreen}{+4.06}) & 27.35 & 23.29 & \textbf{39.32} (\textcolor{posgreen}{+16.67}) \\
\texttt{sing\_col\_sing\_shape} & 23.93 & 31.20 & 28.21 & 33.97 (\textcolor{posgreen}{+10.04}) & 29.06 & 25.21 & \textbf{34.61} (\textcolor{posgreen}{+10.69}) & \textcolor{negred}{18.38} & \textcolor{negred}{18.80} & 28.42 (\textcolor{posgreen}{+4.49}) \\
\bottomrule
\end{tabular}
}
\caption{Comparison of reprompting strategies for \emph{InternVL2-1B} synthetic datasets. Results are averaged across four probe architectures (MLP, Circular-Separately, Logistic-Separately, and Circular-Jointly).}
\label{tab:internvl1b_prompt_comparison}
\end{table*}

\subsection{Comparison of Entropy and Probe-Guided Self-Correction}
\label{appendix:entropy_vs_probe}

The results demonstrate that \textbf{probe-guided self-correction strictly outperforms entropy-based heuristics} across all evaluated datasets. Relying purely on output entropy to trigger self-correction is highly unreliable; it actively degrades model performance (yielding net-negative accuracy deltas) on two of the four datasets (\texttt{diff\_col\_diff\_shape} and \texttt{sing\_col\_sing\_shape}). 

In contrast, the probe-guided approach consistently yields positive gains, achieving a massive \textbf{+16.24\% absolute improvement} on the \texttt{sing\_col\_diff\_shape} split. Mechanistically, the probe-guided method succeeds by accurately targeting and correcting flawed reasoning paths that generic, uncertainty-based entropy metrics fail to effectively address.

\begin{table*}[h!]
\centering
\begin{tabular}{l c cc cc}
\toprule
& & \multicolumn{2}{c}{\textbf{Entropy-Guided}} & \multicolumn{2}{c}{\textbf{Probe-Guided (Ours)}} \\
\cmidrule(lr){3-4} \cmidrule(lr){5-6}
\textbf{Dataset} & \textbf{Base Acc.} & \textbf{Acc.} & \textbf{$\Delta$} & \textbf{Acc.} & \textbf{$\Delta$} \\
\midrule
\texttt{diff\_col\_diff\_shape} & 27.35\% & \textcolor{negred}{25.64\%} & \textcolor{negred}{-1.71\%} & \textbf{34.62\%} & \textcolor{posgreen}{+7.26\%} \\
\texttt{diff\_col\_sing\_shape} & 29.49\% & 31.20\% & \textcolor{posgreen}{+1.71\%} & \textbf{38.03\%} & \textcolor{posgreen}{+8.55\%} \\
\texttt{sing\_col\_diff\_shape} & 22.65\% & 29.06\% & \textcolor{posgreen}{+6.41\%} & \textbf{38.89\%} & \textcolor{posgreen}{+16.24\%} \\
\texttt{sing\_col\_sing\_shape} & 23.93\% & \textcolor{negred}{20.51\%} & \textcolor{negred}{-3.42\%} & \textbf{28.63\%} & \textcolor{posgreen}{+4.70\%} \\
\bottomrule
\end{tabular}
\caption{Comparison of Entropy-guided versus Probe-guided self-correction using the \texttt{counting} prompt strategy.}
\label{tab:entropy_vs_probe_comparison}
\end{table*}

\section{Vision Encoder Probing Results}
\label{appendix:vision_probing}


\begin{figure}[t] 
    \centering
    \begin{subfigure}[b]{1\linewidth} 
        \centering
        \includegraphics[width=\linewidth]{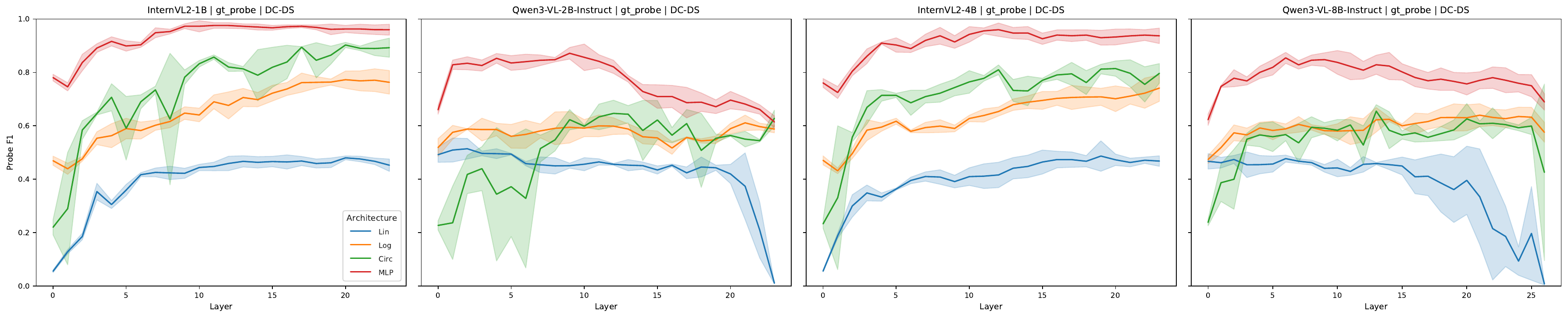}
        \caption{Layer-wise probing F1 scores across vision encoder depths (ground truth objective)}
    \end{subfigure}
    
    \vspace{0.5cm} 

    \begin{subfigure}[b]{1\linewidth} 
        \centering
        \includegraphics[width=\linewidth]{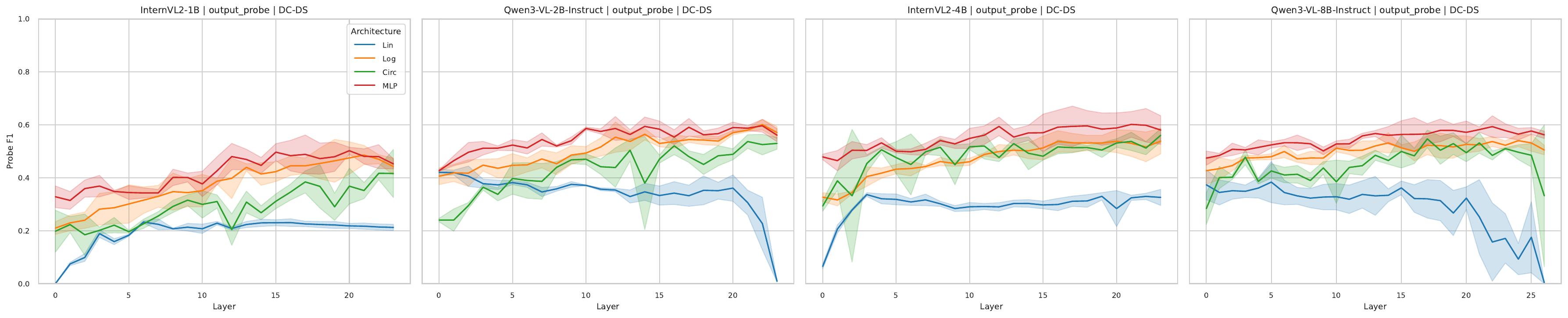}
        \caption{Layer-wise probing F1 score across vision encoder depths (model output objective).}
    \end{subfigure}
    
    \vspace{0.5cm} 
    
    \begin{subfigure}[b]{1\linewidth}
        \centering
        \includegraphics[width=\linewidth]{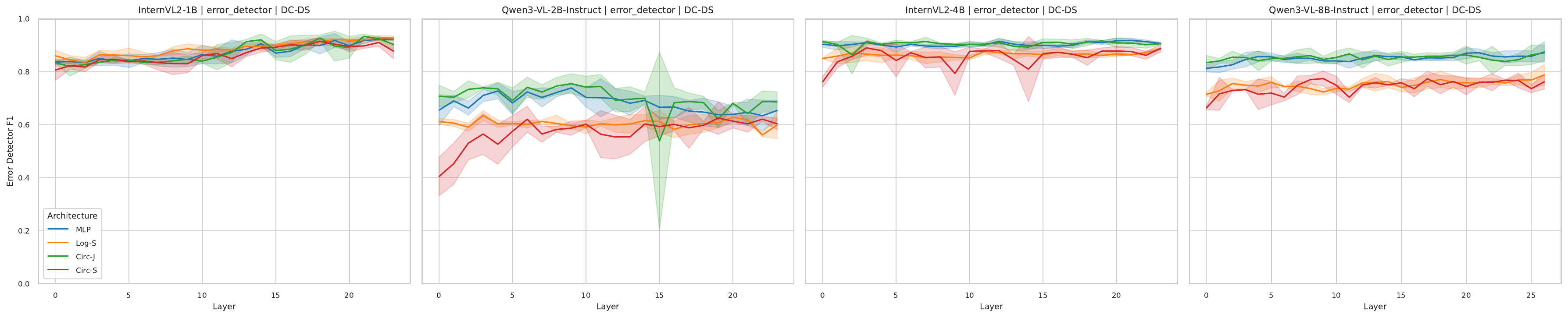}
        \caption{Error detection performance (F1-score) in the vision encoder.}
    \end{subfigure}
    
    \caption{This figure illustrates performance aggregated across models and seeds, where activations for each image are computed as the mean of tokens across all image patches. The top plot shows ground-truth probe F1, measuring feature alignment with the actual object count. The middle plot tracks output-supervised probe F1, assessing alignment with the model's first-pass result. The bottom plot displays error-detector F1, indicating the system's reliability in predicting mistakes to trigger the intervention path. DC-DS: Different Color Different Shape}
    \label{fig:vision_combined_probing_results__diff_col_diff_shape}
\end{figure}

\begin{figure}[t] 
    \centering
    \begin{subfigure}[b]{1\linewidth} 
        \centering
        \includegraphics[width=\linewidth]{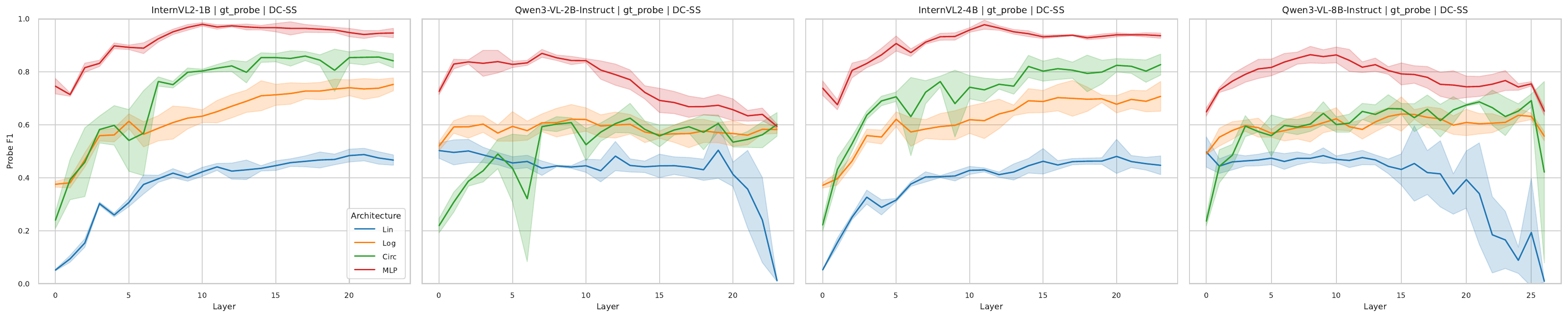}
        \caption{Layer-wise probing F1 scores across vision encoder depths (ground truth objective)}
    \end{subfigure}
    
    \vspace{0.5cm} 

    \begin{subfigure}[b]{1\linewidth} 
        \centering
        \includegraphics[width=\linewidth]{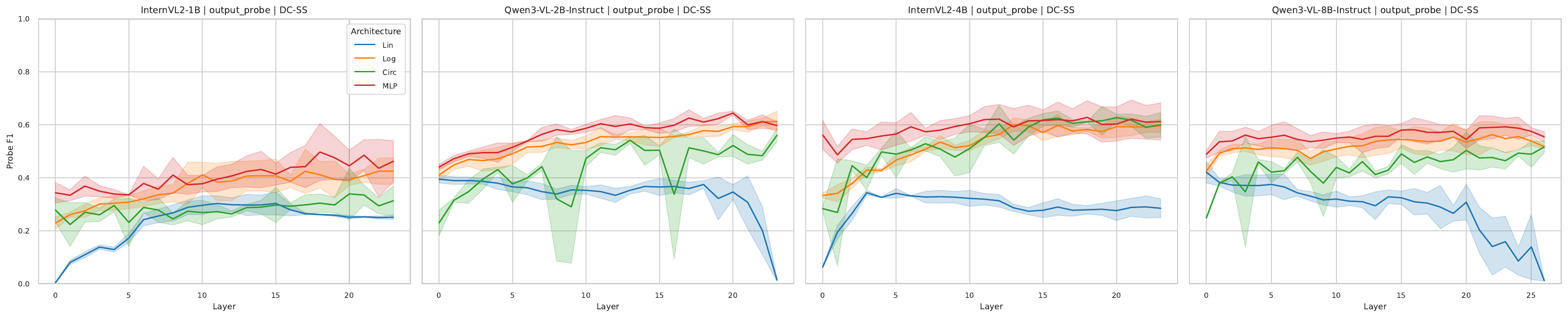}
        \caption{Layer-wise probing F1 score across vision encoder depths (model output objective).}
    \end{subfigure}
    
    \vspace{0.5cm} 
    
    \begin{subfigure}[b]{1\linewidth}
        \centering
        \includegraphics[width=\linewidth]{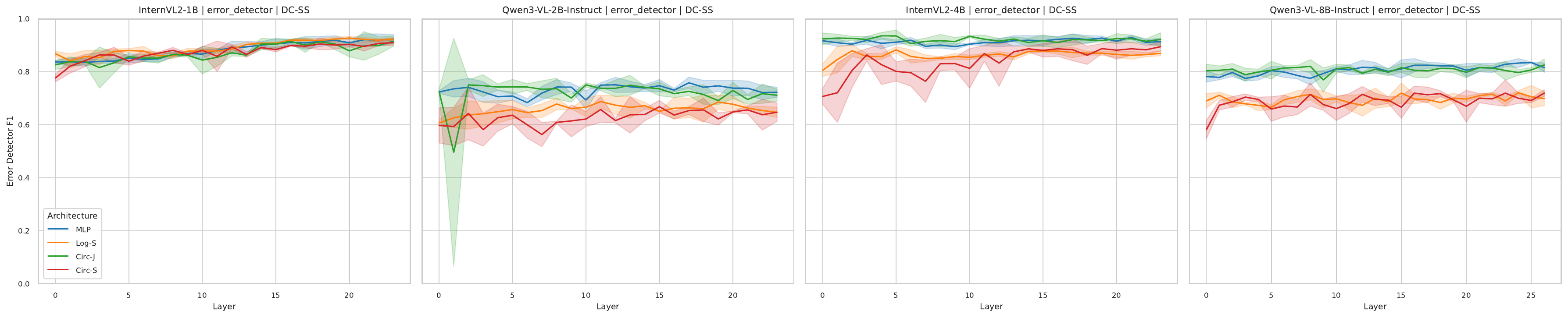}
        \caption{Error detection performance (F1-score) in the vision encoder.}
    \end{subfigure}
    
    \caption{This figure illustrates performance aggregated across models and seeds, where activations for each image are computed as the mean of tokens across all image patches. The top plot shows ground-truth probe F1, measuring feature alignment with the actual object count. The middle plot tracks output-supervised probe F1, assessing alignment with the model's first-pass result. The bottom plot displays error-detector F1, indicating the system's reliability in predicting mistakes to trigger the intervention path. DC-SS: Different Color Single Shape}
    \label{fig:vision_combined_probing_results__diff_col_sing_shape}
\end{figure}

\begin{figure}[t] 
    \centering
    \begin{subfigure}[b]{1\linewidth} 
        \centering
        \includegraphics[width=\linewidth]{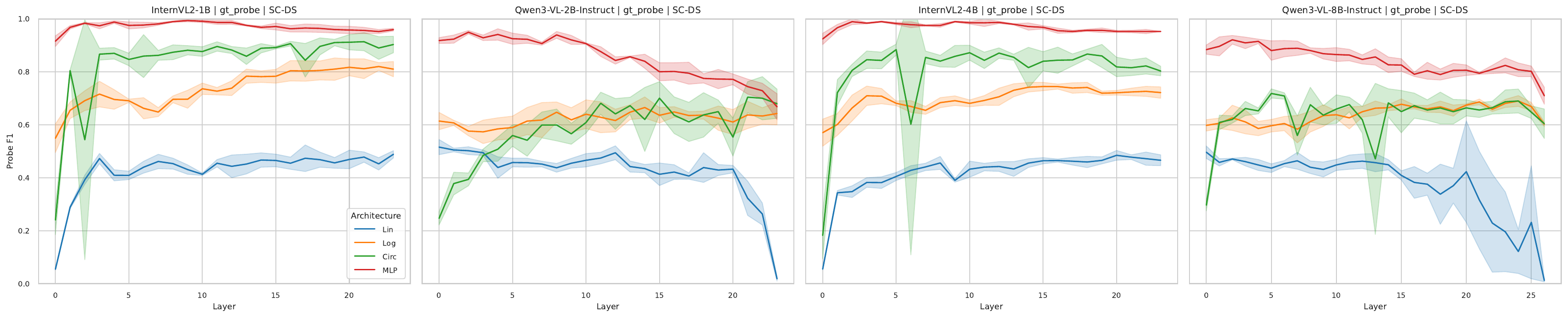}
        \caption{Layer-wise probing F1 scores across vision encoder depths (ground truth objective)}
    \end{subfigure}
    
    \vspace{0.5cm} 

    \begin{subfigure}[b]{1\linewidth} 
        \centering
        \includegraphics[width=\linewidth]{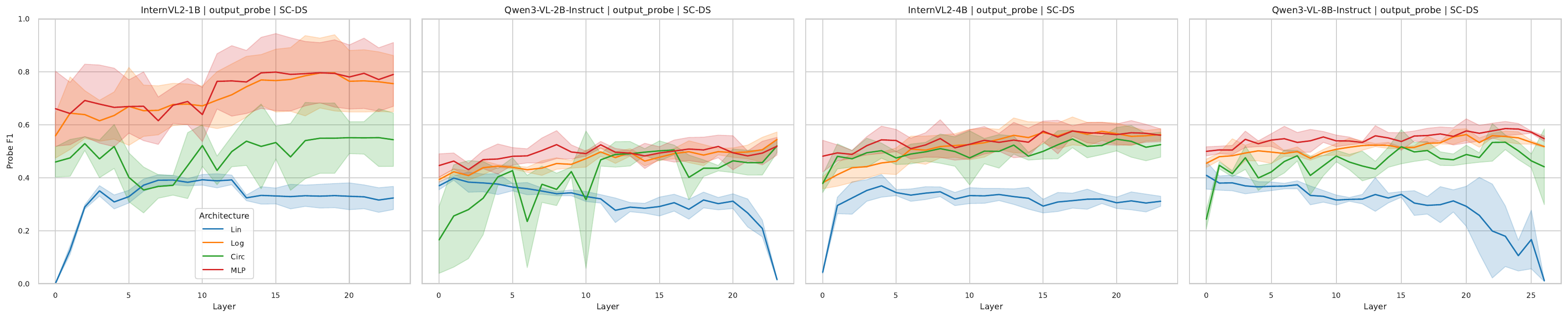}
        \caption{Layer-wise probing F1 score across vision encoder depths (model output objective).}
    \end{subfigure}
    
    \vspace{0.5cm} 
    
    \begin{subfigure}[b]{1\linewidth}
        \centering
        \includegraphics[width=\linewidth]{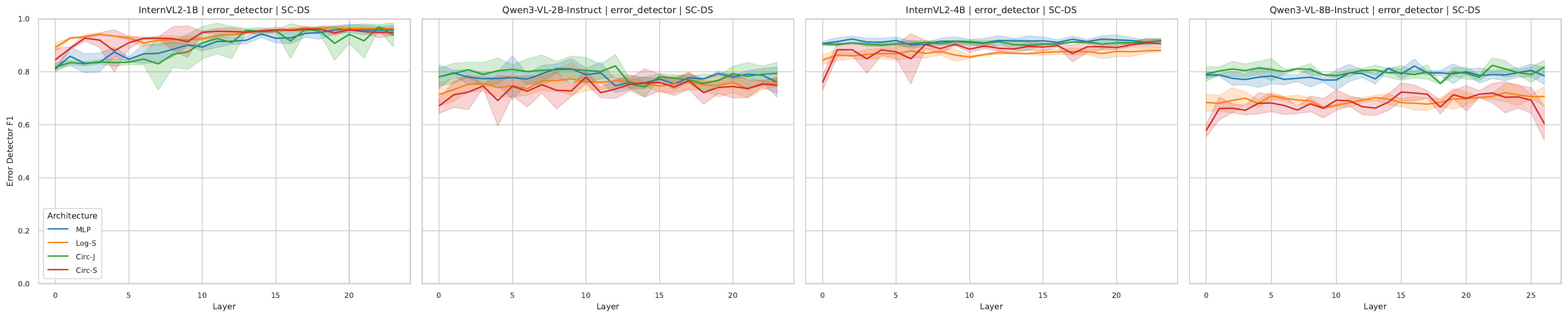}
        \caption{Error detection performance (F1-score) in the vision encoder.}
    \end{subfigure}
    
    \caption{This figure illustrates performance aggregated across models and seeds, where activations for each image are computed as the mean of tokens across all image patches. The top plot shows ground-truth probe F1, measuring feature alignment with the actual object count. The middle plot tracks output-supervised probe F1, assessing alignment with the model's first-pass result. The bottom plot displays error-detector F1, indicating the system's reliability in predicting mistakes to trigger the intervention path. SC-DS: Single Color Different Shape}
    \label{fig:vision_combined_probing_results__sing_col_diff_shape}
\end{figure}

\begin{figure}[t] 
    \centering
    \begin{subfigure}[b]{1\linewidth} 
        \centering
        \includegraphics[width=\linewidth]{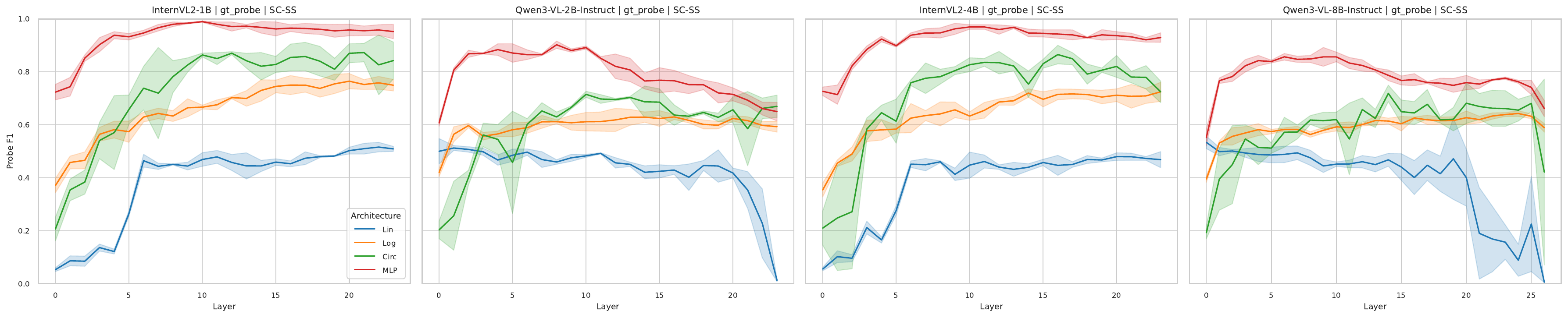}
        \caption{Layer-wise probing F1 scores across vision encoder depths (ground truth objective)}
    \end{subfigure}
    
    \vspace{0.5cm} 

    \begin{subfigure}[b]{1\linewidth} 
        \centering
        \includegraphics[width=\linewidth]{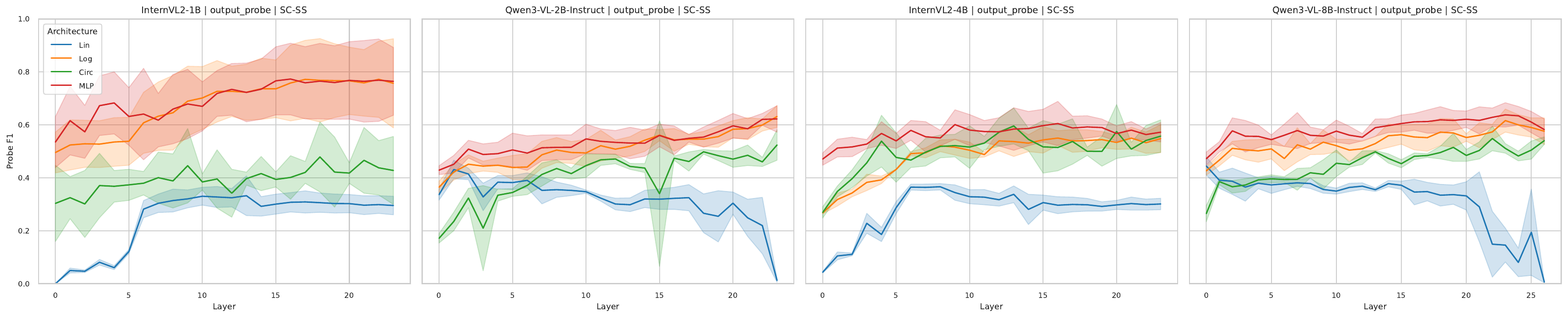}
        \caption{Layer-wise probing F1 score across vision encoder depths (model output objective).}
    \end{subfigure}
    
    \vspace{0.5cm} 
    
    \begin{subfigure}[b]{1\linewidth}
        \centering
        \includegraphics[width=\linewidth]{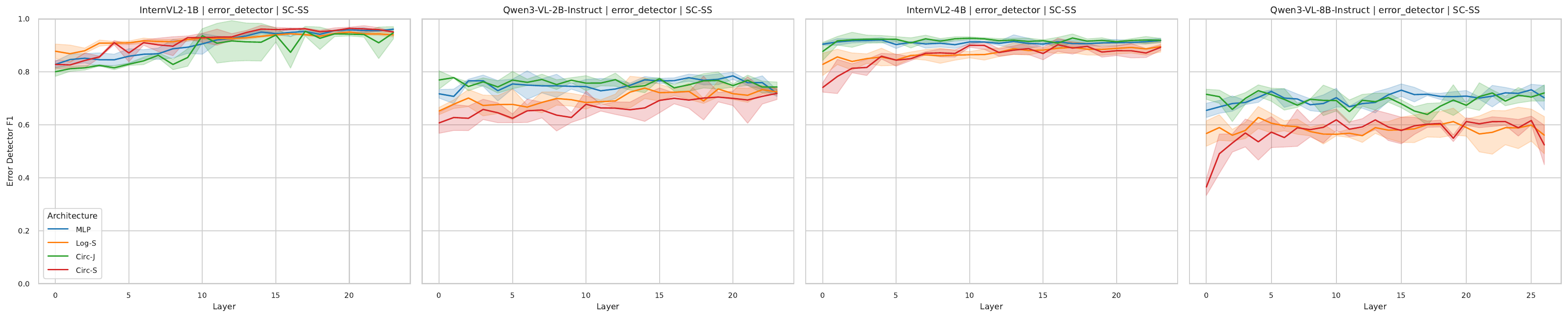}
        \caption{Error detection performance (F1-score) in the vision encoder.}
    \end{subfigure}
    
    \caption{This figure illustrates performance aggregated across models and seeds, where activations for each image are computed as the mean of tokens across all image patches. The top plot shows ground-truth probe F1, measuring feature alignment with the actual object count. The middle plot tracks output-supervised probe F1, assessing alignment with the model's first-pass result. The bottom plot displays error-detector F1, indicating the system's reliability in predicting mistakes to trigger the intervention path. SC-SS: Single Color Single Shape}
    \label{fig:vision_combined_probing_results__sing_col_sing_shape}
\end{figure}

\begin{figure}[t] 
    \centering
    \begin{subfigure}[b]{1\linewidth} 
        \centering
        \includegraphics[width=\linewidth]{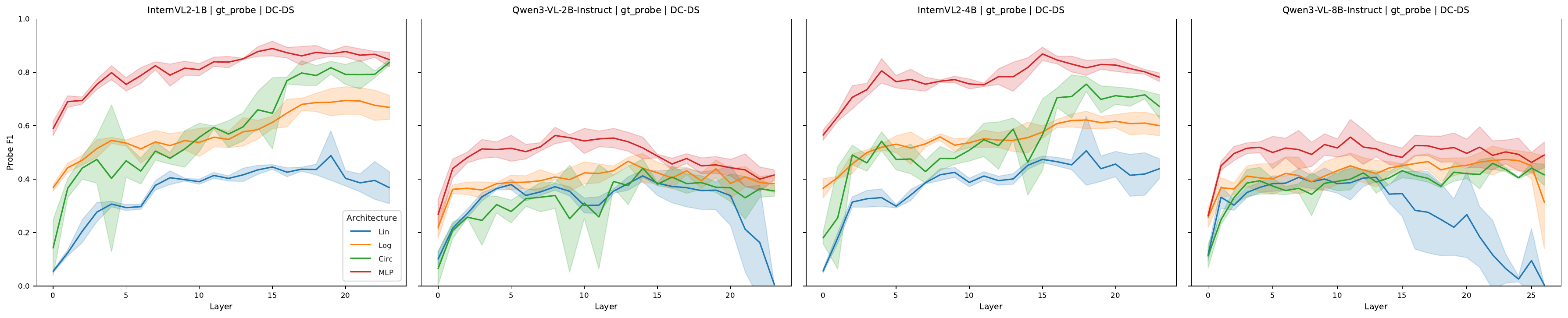}
        \caption{Layer-wise probing F1 scores across vision encoder depths (ground truth objective)}
    \end{subfigure}
    
    \vspace{0.5cm} 

    \begin{subfigure}[b]{1\linewidth} 
        \centering
        \includegraphics[width=\linewidth]{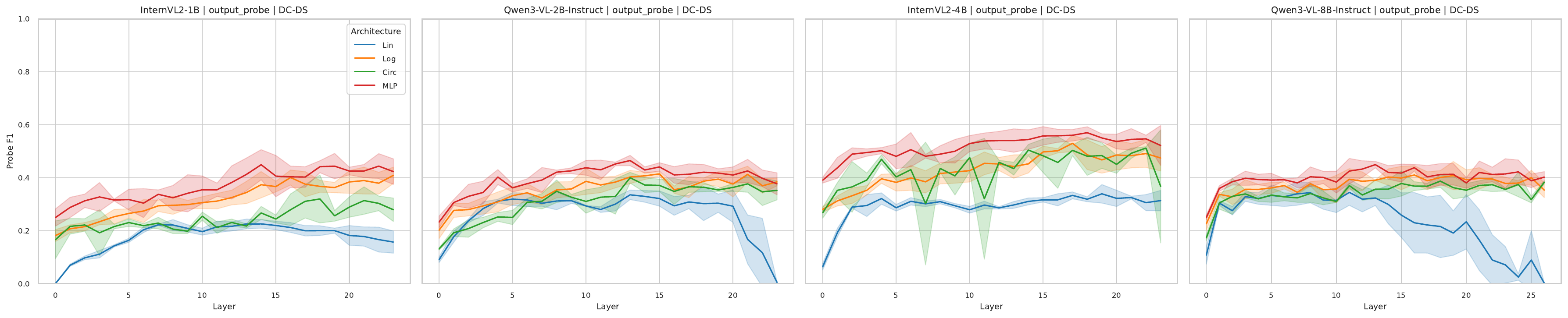}
        \caption{Layer-wise probing F1 score across vision encoder depths (model output objective).}
    \end{subfigure}
    
    \vspace{0.5cm} 
    
    \begin{subfigure}[b]{1\linewidth}
        \centering
        \includegraphics[width=\linewidth]{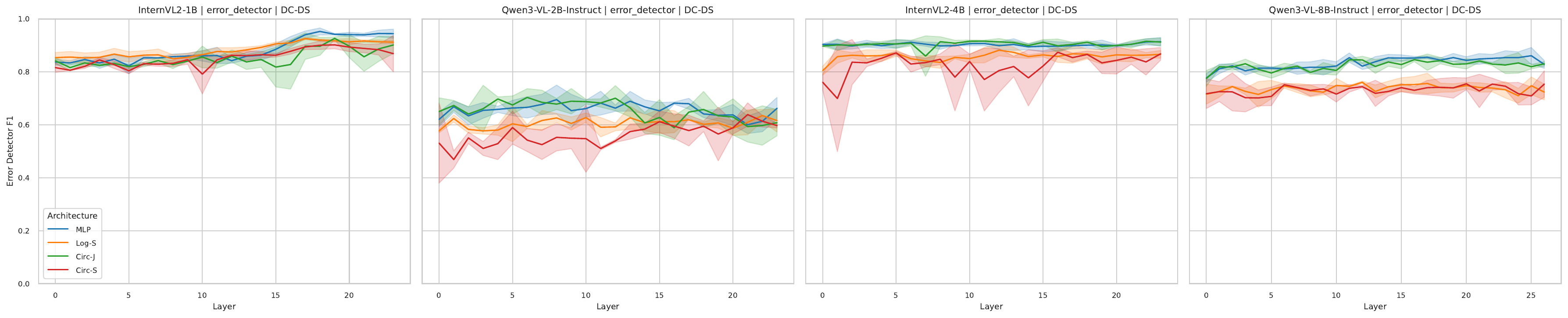}
        \caption{Error detection performance (F1-score) in the vision encoder.}
    \end{subfigure}
    
    \caption{This figure illustrates performance aggregated across models and seeds, where activations for each image are extracted from the last token of the vision encoder. The top plot shows ground-truth probe F1, measuring feature alignment with the actual object count. The middle plot tracks output-supervised probe F1, assessing alignment with the model's first-pass result. The bottom plot displays error-detector F1, indicating the system's reliability in predicting mistakes to trigger the intervention path. DC-DS: Different Color Different Shape}
    \label{fig:vision_combined_probing_results__diff_col_diff_shape_last}
\end{figure}

\begin{figure}[t] 
    \centering
    \begin{subfigure}[b]{1\linewidth} 
        \centering
        \includegraphics[width=\linewidth]{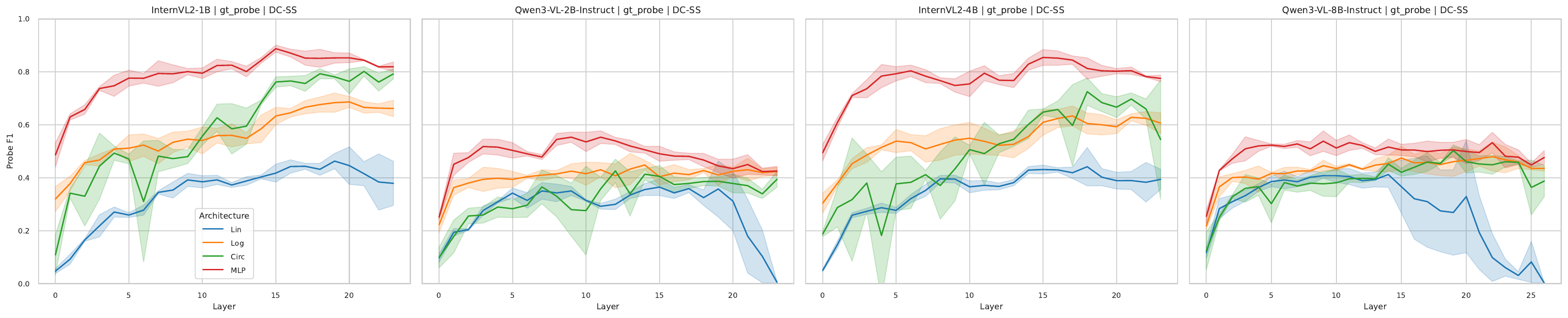}
        \caption{Layer-wise probing F1 scores across vision encoder depths (ground truth objective)}
    \end{subfigure}
    
    \vspace{0.5cm} 

    \begin{subfigure}[b]{1\linewidth} 
        \centering
        \includegraphics[width=\linewidth]{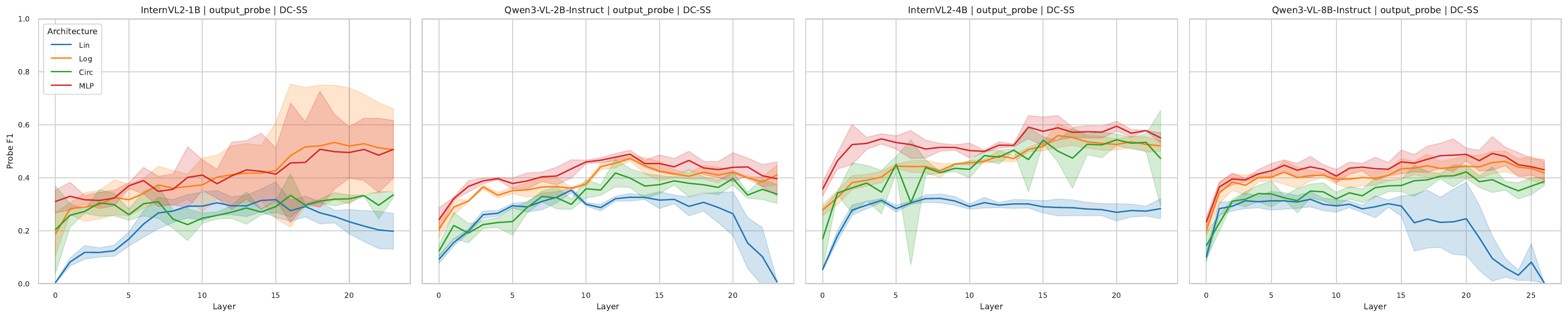}
        \caption{Layer-wise probing F1 score across vision encoder depths (model output objective).}
    \end{subfigure}
    
    \vspace{0.5cm} 
    
    \begin{subfigure}[b]{1\linewidth}
        \centering
        \includegraphics[width=\linewidth]{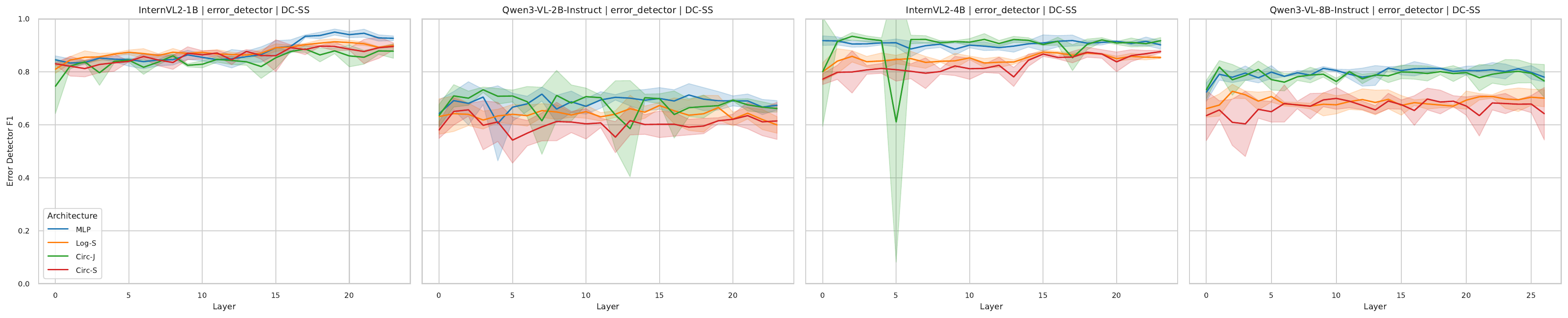}
        \caption{Error detection performance (F1-score) in the vision encoder.}
    \end{subfigure}
    
    \caption{This figure illustrates performance aggregated across models and seeds, where activations for each image are extracted from the last token of the vision encoder. The top plot shows ground-truth probe F1, measuring feature alignment with the actual object count. The middle plot tracks output-supervised probe F1, assessing alignment with the model's first-pass result. The bottom plot displays error-detector F1, indicating the system's reliability in predicting mistakes to trigger the intervention path. DC-SS: Different Color Single Shape}
    \label{fig:vision_combined_probing_results__diff_col_sing_shape_last}
\end{figure}

\begin{figure}[t] 
    \centering
    \begin{subfigure}[b]{1\linewidth} 
        \centering
        \includegraphics[width=\linewidth]{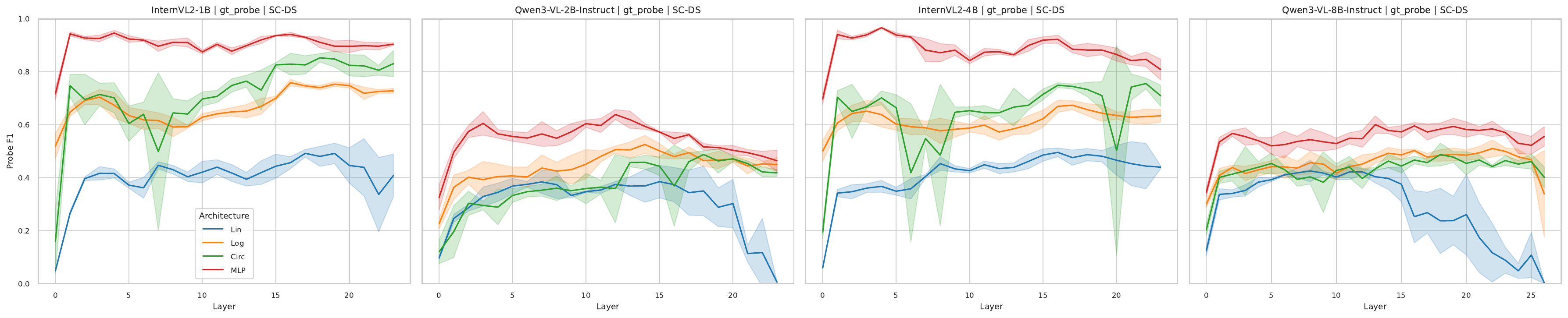}
        \caption{Layer-wise probing F1 scores across vision encoder depths (ground truth objective)}
    \end{subfigure}
    
    \vspace{0.5cm} 

    \begin{subfigure}[b]{1\linewidth} 
        \centering
        \includegraphics[width=\linewidth]{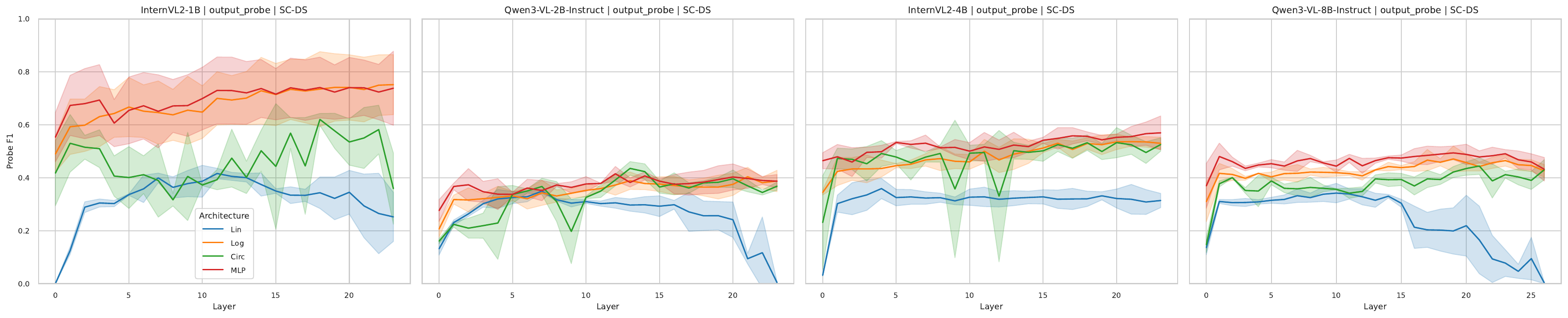}
        \caption{Layer-wise probing F1 score across vision encoder depths (model output objective).}
    \end{subfigure}
    
    \vspace{0.5cm} 
    
    \begin{subfigure}[b]{1\linewidth}
        \centering
        \includegraphics[width=\linewidth]{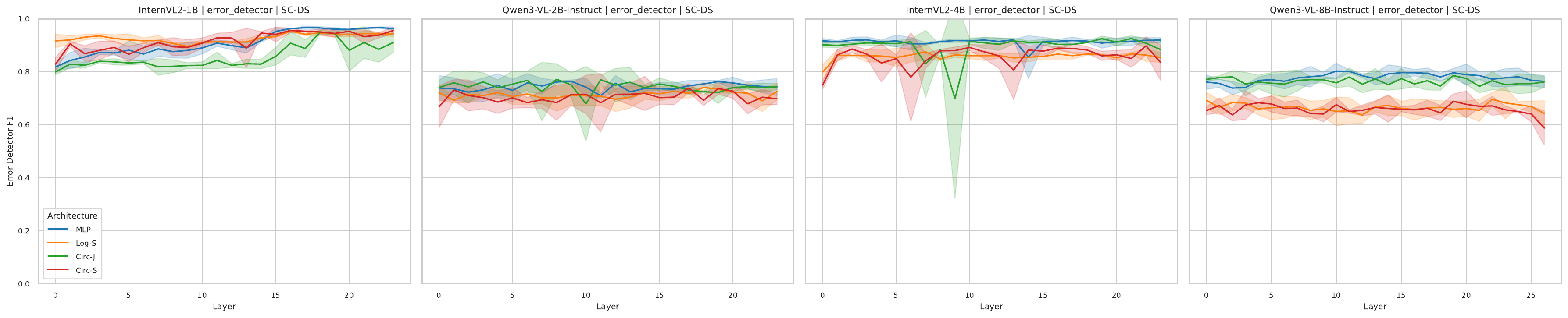}
        \caption{Error detection performance (F1-score) in the vision encoder.}
    \end{subfigure}
    
    \caption{This figure illustrates performance aggregated across models and seeds, where activations for each image are extracted from the last token of the vision encoder. The top plot shows ground-truth probe F1, measuring feature alignment with the actual object count. The middle plot tracks output-supervised probe F1, assessing alignment with the model's first-pass result. The bottom plot displays error-detector F1, indicating the system's reliability in predicting mistakes to trigger the intervention path. SC-DS: Single Color Different Shape}
    \label{fig:vision_combined_probing_results__sing_col_diff_shape_last}
\end{figure}

\begin{figure}[t] 
    \centering
    \begin{subfigure}[b]{1\linewidth} 
        \centering
        \includegraphics[width=\linewidth]{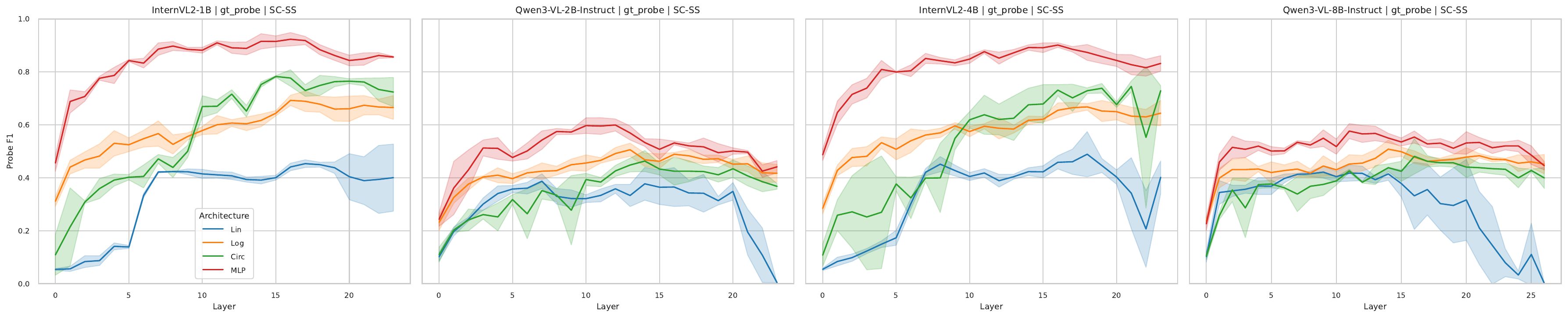}
        \caption{Layer-wise probing F1 scores across vision encoder depths (ground truth objective)}
    \end{subfigure}
    
    \vspace{0.5cm} 

    \begin{subfigure}[b]{1\linewidth} 
        \centering
        \includegraphics[width=\linewidth]{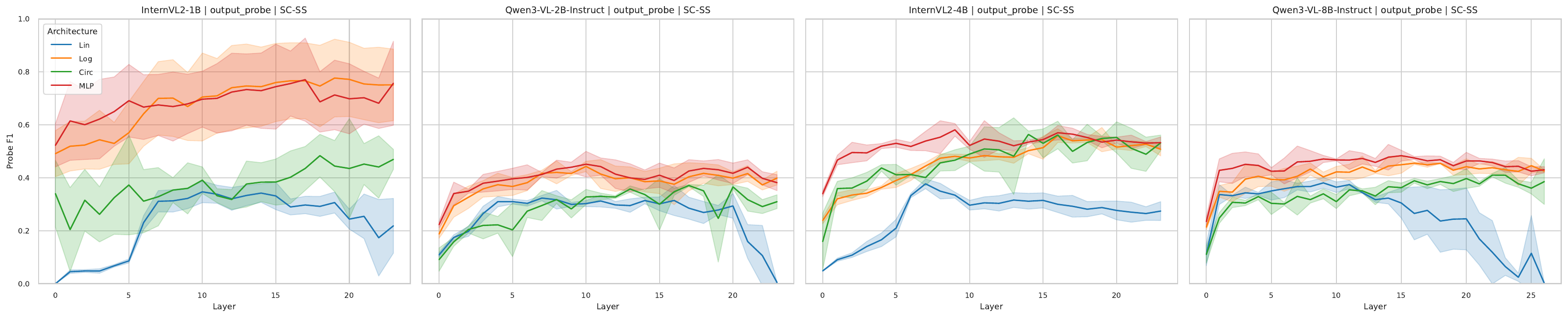}
        \caption{Layer-wise probing F1 score across vision encoder depths (model output objective).}
    \end{subfigure}
    
    \vspace{0.5cm} 
    
    \begin{subfigure}[b]{1\linewidth}
        \centering
        \includegraphics[width=\linewidth]{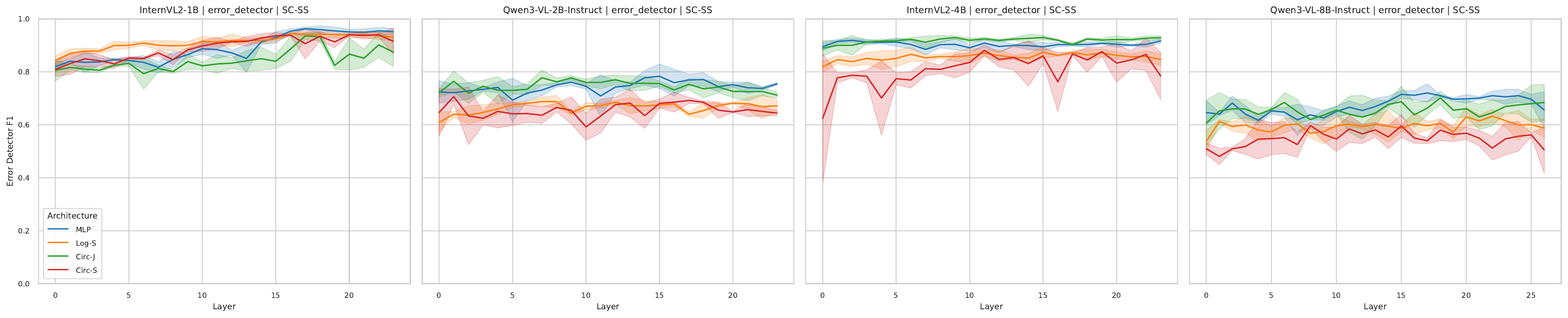}
        \caption{Error detection performance (F1-score) in the vision encoder.}
    \end{subfigure}
    
    \caption{This figure illustrates performance aggregated across models and seeds, where activations for each image are extracted from the last token of the vision encoder. The top plot shows ground-truth probe F1, measuring feature alignment with the actual object count. The middle plot tracks output-supervised probe F1, assessing alignment with the model's first-pass result. The bottom plot displays error-detector F1, indicating the system's reliability in predicting mistakes to trigger the intervention path. SC-SS: Single Color Single Shape}
    \label{fig:vision_combined_probing_results__sing_col_sing_shape_last}
\end{figure}

Testing was conducted across two distinct settings: probes were trained on activations extracted from the last vision token of the VLM vision encoder (results shown in Figures ~\ref{fig:vision_combined_probing_results__diff_col_diff_shape_last},~\ref{fig:vision_combined_probing_results__diff_col_sing_shape_last},~\ref{fig:vision_combined_probing_results__sing_col_diff_shape_last} and ~\ref{fig:vision_combined_probing_results__sing_col_sing_shape_last} and on the mean activations across all tokens in each image (results shown in Figures ~\ref{fig:vision_combined_probing_results__diff_col_diff_shape},~\ref{fig:vision_combined_probing_results__diff_col_sing_shape},~\ref{fig:vision_combined_probing_results__sing_col_diff_shape} and ~\ref{fig:vision_combined_probing_results__sing_col_sing_shape}). 

The results of the layered analysis of the DC-DS, DC-SS, SC-DS and SC-SS data sets reveal a persistent mechanistic gap between the internal perception of the model and its final verbalized result. Specifically, multilayer perceptrons (MLPs) reach peak performance very early - often by the 5th or 10th layer and maintain this plateau until the final layers. This suggests that the "true" answer is successfully extracted from the visual input and represented within the model early in the forward pass. The model "knows" the answer long before it reaches the final output layers, where the decision to verbalize it is made. It is also worth noting that the ground-truth probes of Qwen-3-VL models begin to fall after the middle layers, which is not observed in the InternVL-2 models. Unlike ground-truth probes, which plateau early in the process, output probes typically reach peak F1 values only in the final layers of the model, indicating that the representational subspace governing actual count generation is fully formed only at the end of the processing pipeline. This trend persists across varying visual representation complexities, although the performance of these samples is generally lower than that of their ground-truth counterparts.

Furthermore, high F1-score values obtained by error detection algorithms trained on vision encoder activations are likely due to a spurious correlation between scene complexity and the model's error rate, rather than genuine error detection. Because the vision language model (VLM) makes more errors on images with a large number of objects, which also tend to be visually denser and cluttered, the vision encoder activations for "complex" scenes (many objects) and "simple" scenes (few objects) systematically differ. The error detection algorithm exploits this: it learns to predict "error" for activations corresponding to complex scenes and "correct" for simple ones, effectively acting as a scene complexity classifier rather than an error detector. Because complexity and error coexist in the dataset, this leads to inflated F1-score varies that do not reflect the model's true ability to determine whether the model succeeded or failed on any given image.

Comparative analysis shows that probes trained on average activation values consistently outperform probes restricted to the last visual token across all four levels of visual complexity. This performance advantage likely stems from the fact that the critical spatial information required for accurate counting is not concentrated in a single patch or token, but is distributed across the entire image grid. By aggregating features across all patches, averaged representations allow for a more holistic representation of the scene, reducing the information loss inherent in extracting individual tokens and providing a more robust signal for probes to identify internally represented values.

\end{document}